\documentclass[12pt]{third-rep}
\usepackage[T1]{fontenc}
\usepackage{kpfonts}
\usepackage[scaled=0.85]{beramono}
\usepackage[margin=0.9in]{geometry}
\usepackage{amsmath}

\usepackage{subcaption}
\usepackage{multirow}
\usepackage{graphicx}
\usepackage[table,xcdraw]{xcolor}
\usepackage{multirow}
\usepackage{colortbl}
\usepackage{hhline}
 \usepackage{array}
 \usepackage{multirow}
 \usepackage{ragged2e}
 \usepackage{colortbl}
 \usepackage{epigraph}
 \usepackage{amsmath}
\usepackage{mathtools}
\usepackage{multirow}
\usepackage{longtable}

\title{Graph Neural Networks for Cancer Data Integration}
\author{Teodora Reu}
\supervisor{Shapiro Jonathan}
\reportyear{2021-2022}

\abstractfile{abstract}
\thanksfile{dedic}

\usepackage{fancyhdr}
\usepackage{pslatex}

\usepackage{listings}
\usepackage{csvsimple}
\usepackage{tgbonum}
\usepackage[scaled=0.85]{beramono}
\usepackage{kpfonts}

\begin{document}
\includegraphics[width=0.4\textwidth]{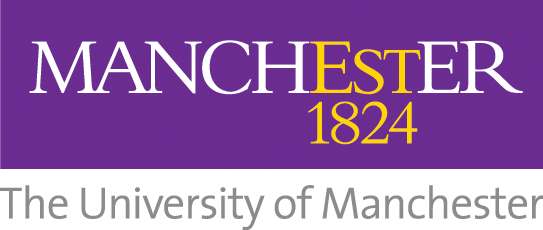}
\dotitleandabstract

\clearpage
\begin{center}
    \thispagestyle{empty}
    \vspace*{\fill}
    \textit{
    I dedicate this project to Nicolae Reu, my grandpa, who asked me at the age of seven to compute $\pi$ with a string and a ruler, and to my grandma's - Săndina Haja - memory.}
    \vspace*{\fill}
\end{center}
\clearpage

\tableofcontents
\listoffigures
\listoftables
\chapter{Introduction}
\label{cha:intro}

\epigraph{Someday my paintings will be hanging in the Louvre.}{Lust for life, Irving Stone}

Biomedicine is the field of science that takes the research findings of medical science and applies them in clinical practice. For a long time, it has based its conclusions on the correlation analysis of various events. For example, when a patient exhibits symptoms A, B, and C, there is a certain probability that they have disease X. Medical practitioners learn the correlations between symptoms and conditions, with the savviest easily finding associations leading to the correct diagnosis i.e. based on the inputs they can generate an output with high confidence. This can be pictured as having a space with N dimensions resembling all possible symptoms, and a doctor embedding the patients as data points in this space. Thus, if a new subject resembles the symptoms of previously diagnosed patients, the doctor can produce a list of possible conditions based on the similarity to the previous patients. In technical terms, the output is based on the proximity to the already classified data points.

This represents a viable way to understand biomedical data, with the addition that over the past decade, a wide range of new data types describing patients has been collected, ranging from graphical data such as X-ray or MRI scans to cellular quantities of proteins or genes and concentrations of minerals in the blood, among many others. These collections contain large volumes of data of various shapes and sources, both sparse and dense, encompassing categorical and continuous values. Despite these recent efforts, medical experts still face difficulties in making diagnoses given the broad set of possible symptoms, hence the demand for AI models which can learn the correlations that the human brain might overlook.

In the next sections, we will introduce the efforts in collecting cell-related data (referred to as multi-omic or multi-genomic) from various organisms, followed by the existing research aiming to better understand these data, i.e. by further categorizing the types of diseases. Finally, we will present the case for why unsupervised learning with Graph Neural Networks has the potential to show promising results and what these would look like, and then, the goal of this project.
\section{Efforts on collecting data}
On the past decade, many projects provided comprehensive catalogues of cells coming from organisms with certain diseases. By using cutting-edge single-cell and spatial genomics alongside computational techniques, the Human Cell Atlas \cite{regev2017science} researchers revealed that approximately 20,000 genes in a individual cell can be switched on, creating a unique identity for each cell. This has been studied on more than 24.6 million cells coming from multiple organs such as: Brain, Pancreas, Blood, Kidney etc. Another notable cell atlas is Brain Research through Advancing Innovative Techniques (BRAIN) \cite{ecker2017brain} which studies single cells during health and disease periods. Other examples of such atlases are: Cell Atlas of Worm \cite{diaz2020single}, Fly Cell Atlas \cite{li2022fly}, and Tabula Muris \cite{tabula2018single}. 

Among the consortia representing large collections of cancer data, the most notable are The Cancer Genome Atlas (TCGA), and the Molecular Taxonomy
of Breast Cancer International Consortium (\textbf{METABRIC}), which describes 2000 breast cancer patients by using multi-omic and clinical data. 
\section{Efforts on integrating heterogeneous cancer data}
Multiple sources such as \cite{biswas2020artificial} and \cite{yarmishyn2015long} argue that in order to give better treatment to cancer patients the integration of multi-omic (cell mRNA expression, DNA methylation) and clinical data would be preferable. They suggest that patients with the same cancer type should be given different treatments based on their integrated results, thus leading to a further sub-categorisation of the usual types of cancer.

This can be achieved by clustering patient data points based on their multi-omic and clinical results, many models that attempt integration of such data have been tried over the last decade. Since, the sub-categories of cancer types are yet unknown, unsupervised models were trained on patients data to underline different categories of the same cancer type. For example, knowing that patients from a found sub-category received two types of treatments, with a higher survival rate for one of the treatments, could help practitioners decide which treatment works best for that sub-category of patients.

The different integration types can be observed in the table below.

\begin{table}[h]
\centering
\small
\scalebox{0.65}{%
\begin{tabular}{clll}
\multicolumn{1}{l}{Integration Type} & Method & Comments & Citations \\ \hline
Early &
  \begin{tabular}[c]{@{}l@{}}Plain concatenation of every dataset into\\ a single, large matrix, and then apply models\end{tabular} &
  \begin{tabular}[c]{@{}l@{}}- The resulting dataset is noisier \\ and high dimensional,  which \\makes learning more difficult\\ - Overlooks omics with \\ smaller feature dimension\\ - Ignores the specific data \\ distribution of each omic\end{tabular} &
  \begin{tabular}[c]{@{}l@{}}\cite{xie2019group}\\ \cite{chaudhary2018deep}\end{tabular} \\ \hline
Mixed &
  \begin{tabular}[c]{@{}l@{}} Concatenation of lower latent space representations\\ takes place in the middle of the learning process.\\\end{tabular} &
  \begin{tabular}[c]{@{}l@{}}- Addresses the short comings \\ of the early integration types\\ - Some of the proposed models are:\\ Kernel Learning, Graph Based, ANN\end{tabular} &
  \begin{tabular}[c]{@{}l@{}}\cite{he2021integrating}\\ \cite{wang2017visualization}\\ \cite{wang2014similarity}\\ \cite{ma2018affinity}\\ \cite{lee2020heterogeneous}\\ \cite{zitnik2018modeling}\end{tabular} \\ \hline
Late &
  \begin{tabular}[c]{@{}l@{}}Applies the model on the datasets independently,\\ followed by an aggregation function over the output \\for cancer prognosis \end{tabular} &
  \begin{tabular}[c]{@{}l@{}}- Cannot capture the inter-omic interactions \\ - The models do not share information about \\the learned features\\\end{tabular} &
  \begin{tabular}[c]{@{}l@{}}\cite{sun2018multimodal}\\ \cite{wang2020moronet}\end{tabular} \\ \hline
Hierarchical &
  \begin{tabular}[c]{@{}l@{}}    Generates a different representation for\\ each genomic in part, which are concatenated\\ and used to train an encoding model\end{tabular} &
  \begin{tabular}[c]{@{}l@{}} - Some of the proposed models are: iBag\\(integration Bayesian  analysis of genomics),\\ LRMs(linear regulatory modules), \\ARMI(Assisted Robust Marker Identification)\end{tabular} &
  \begin{tabular}[c]{@{}l@{}}\\ \cite{wang2013ibag}\\ \cite{zhu2016integrating}\\ \cite{chai2017analysis}\end{tabular} \\ \hline
\end{tabular}%
}
\label{tab:methods}
\caption{Table of state-of-the-art cancer data integration approaches inspired by the review \cite{PICARD20213735}}
\end{table}
Next subsections show relevant projects to what this project's goal will be. 
\subsubsection{Autoencoders}
In \cite{chaudhary2018deep}, they used unsupervised and supervised learning on two subgroups with significant survival differences. Extending the study to multiple types of cohorts of varying ethnicity has helped identify 10 consensus driving genes by association with patients survival \cite{chaudhary2019multimodal}. 

In \cite{xu2019hierarchical}, the stacked autoencoder was used on each modality, and then the extracted representation represented the input to another autoencoder. Finally, a supervised method was used to evaluate the quality of the lower space representations.

In \cite{SimidjievskiEtAl2019} the authors use several variational autoencoders architectures to integrate multi-omic and clinical data. They evaluate their integrative approaches by combining pairs of modalities, and by testing if their lower latent space where sensitive enough to resemble certain cancer sub-types.

\subsubsection{Similarity network fusion}
In \cite{wang2014similarity} the way Similarity Network Fusion (SNF) constructed networks using multi-genomic data is related to this project, since that it attempts learning on constructed graph on multi-omic and clinical data. Given two or more types of data for the same objects (e.g. patients), SNF will firstly create one individual network for each modality, by using a similarity patient-to-patient similarity measure. After that, a network fusion step will take place, which uses a nonlinear method based on message passing theory \cite{monti2003consensus} applied iteratively to the two networks. After a few iterations, SNF converges to a single network. Their method is robust and has a lot of the hyper-parameter settings. The advantage of their approach is that weak similarities will disappear in time, and strong similarities will become even stronger over time. 

In the next section, we will present promising models which perform unsupervised learning over graph shaped like data, and get very good results on a clustering task.

\section{Unsupervised Graph Neural Networks}
Among many unsupervised Graph Neural Networks, some of the notable ones are Autoencoder \cite{kipf2016variational} and Deep Graph Infomax \cite{velickovic2019deep}.

\textbf{Variational Graph Autoencoders} (VGAEs) have shown very promising results summarizing graph's structure, by leveraging very good results on link prediction task. The experiment is based on encoding the graph's structure (with randomly missing edges) in lower space representation, attempt reconstruction of the adjacency matrix, and then compare the reproduced adjacency matrix, with the complete input adjacency matrix. They compared their models against \textit{spectral clustering} and against \textit{DeepWalk}, and got significantly better results on Cora, Citeseer, Pubmed \cite{sen2008collective}.

Another, very promising unsupervised Graph Neural Network is  \textbf{Deep Graph Infomax} (DGIs), showing very promising results on getting lower space representations on graph shaped datasets. The implicit Deep Infomax even obtains better lower latent space representations on datasets such as CIFAR10, Tiny ImageNet then variational Autoencoders \cite{hjelm2018learning}. The Deep \textbf{Graph} Infomax obtains amazing results on datasets such as Cora, Citesse, Pubmed, Reddit and PPI. In this case the unsupervised DGI gives better results even than supervised learning on a classification task (for supervised learning models use in additions the labels of the object, in unsupervised learning the labels are never seen). The DGI shows itself as a very promising unsupervised model architecture.
\section{Goal of the project}

This project leverages integrative unsupervised learning on Graph Neural Networks, such as VGAEs and DGIs, and aims to obtain competitive results in line with state-of-the-art for cancer data integration \cite{SimidjievskiEtAl2019}. The ultimate goal is to construct robust graphs comprising patient-to-patient relations based on the available data modalities. This can be achieved by  training models in an unsupervised fashion and attempting to combine data sets at different stages throughout the learning process, followed by generating lower-latent space representations. In order to confirm that the resulting embeddings are sensitive enough to underline various sub-types of cancer, we will assess their quality by performing a classification task with a benchmark method such as Naive Bayes on the already remarked sub-types of the breast cancer in the METABRIC dataset. 

A successful project will be characterised by building this novel Machine Learning pipeline - from data source integration and patient-relationship graph construction, to designing models learning lower-dimensional representations which would improve the performance metrics obtained on classification tasks. Ideally, these lower-latent space embeddings will resemble new clusters of data points leading to the discovery of new sub-categories of breast cancer that would help medical practitioners in offering accurate treatments to patients. 

This paper will discuss in depth the following:
\begin{itemize}
    \item Datasets along with each modality in part, the label classes, and the construction of a synthetic dataset which will be used to judge the quality of the proposed models
    \item Artificial neural networks, from Multilayer Perceptron to Autoencoders and Deep Graph Infomax, in order to build knowledge over the models used for unsupervised learning
    \item The recreation of two state-of-the-art models that is useful as a benchmark for the evaluation of the novel models
    \item Building a graph learning pipeline on a non-graph shaped data
    \item Novel models that attempt integration and correctly evaluate the lower latent space embeddings
\end{itemize}

\chapter{Datasets}
\section{METABRIC Dataset}
The METABRIC project is a joint English-Canadian effort to classify breast tumours based on numerous genomic, transcriptomic, and imaging data types collected from over 2000 patient samples \cite{curtis2012genomic}. This data collection is one of the most comprehensive worldwide studies of breast cancer ever undertaken. Similarly to \cite{SimidjievskiEtAl2019}, we will conduct integrative experiments on \textbf{CNAs}, \textbf{mRNA} and \textbf{clinical} data defined below.

\begin{figure}[h!]
    \small
    \centering
    \includegraphics[scale=0.35]{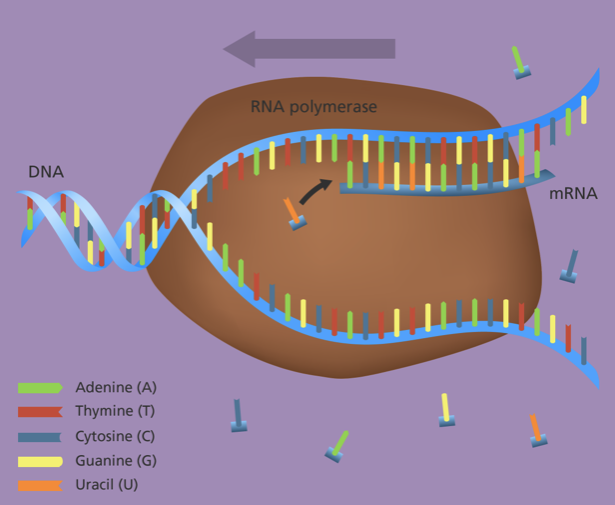}
\caption{Process of transcription borrowed from ``\textit{Genome Research Limited}}"
    \label{fig:mRNA}
\end{figure}

The work in \cite{genome2021your} proposes that gene expression is the process by which instructions in our DNA (\textbf{d}eoxyribo\textbf{n}ucleic \textbf{a}cid) are covered into a functional product, such as a protein. Gene expressions are tightly related to cell responses to changing environments. The process of getting or coping genes chains out of DNA, through messenger-RNA (messenger-ribonucleic acid, or \textbf{mRNA}), is called transcription. RNA is a chemical structure with similar properties as the DNA, with the difference that while DNA has two strands, RNA has only one strand, and instead of the base thymine (T), RNA has a base called uracil (U). The secret of this process is that, if one knows one strand of mRNA, they can guess the other half, and this is because bases come in pairs. For example, if we have a strand \textbf{ACUGU} in a mRNA the other half will be \textbf{TGACA} (because we have Guanine (G) and Cytosine (C) pairs and then Thymine (T) (or Uracil (U) if mRNA) and Adenine (A)). The key property of DNA is the complementarity of its two strands, which allows for accurate replication (DNA to DNA) and information transfer DNA to RNA). This can be easily be seen in Figure \ref{fig:mRNA}. 

\cite{zeira2020copy} describes that an evolutionary process in which somatic (non-multiplicative cells, so neither ovule, sperm) mutations that accumulate in a population of tumor cells result in cancer development. \textbf{Copy number aberrations (CNAs)}, which are the deletion or amplification of large genomic regions, are a type of somatic mutation common to many types of cancer, including breast cancer. CNAs are classified into multiple types and can cover a broad range of sizes, from thousands of kilobases to entire chromosomes and even chromosome arms. A critical role played by CNAs is in driving the development of cancer, and thus the characterization of these events is crucial in the diagnosis, prognosis and treatment of diseases. Furthermore, CNAs act as an important point of reference for reconstructing the evolution of tumors. Although somatic CNAs are the dominant feature discovered in sporadic breast cancer cases, the elucidation of driving events in tumorigenesis is hampered by the large variety of random, non-pathogenic passenger alterations and copy number variants.

METABRIC consists of 1980 breast-cancer patients split in groups based on two immunohistochemistry sub-types, ER+ and ER-, 6 intrinsic gene-expression subtypes (PAM50) \cite{prat12and}, 10 Integrative Clusters (IC10)\cite{curtis2012genomic}, and two groups based on Distance Relapse (the cancer metastasised to another organ after initial treatment or not). 

The dataset which we are going to use is the one already pre-processed by \cite{SimidjievskiEtAl2019}, because the is already split in five-fold cross evaluation, for each labels class, in order to obtain proportional number of object which same class allover the folds. CNA modality has been processed as well, for it's feature to come from a Bernoulli distribution, and the clinical data has been filtered through one-hot-encoding process.
\section{Synthetic Dataset}
When developing novel models on top of complex datasets such as METABRIC, it is hard to segregate the source of any errors or results that fail to meet expectations due to the multitude of stages in the learning pipeline: the data integration and graph building cycle, the model training or the classification task. Thus, we will leverage a testing methodology popular in literature to help us point out any errors or inconsistencies, either from a data, architecture design or implementation perspective. Specifically, we will generate a synthetic dataset coming from a Gaussian distribution; this is advantageous because edges can be predefined based on the labels of the artificial data points, and we are also in control of dataset characteristics such as homophily levels.

\begin{figure}[h!]
    \centering
    \subfloat[\centering Good choice of $\vec{\sigma}$]{{\includegraphics[width=5cm]{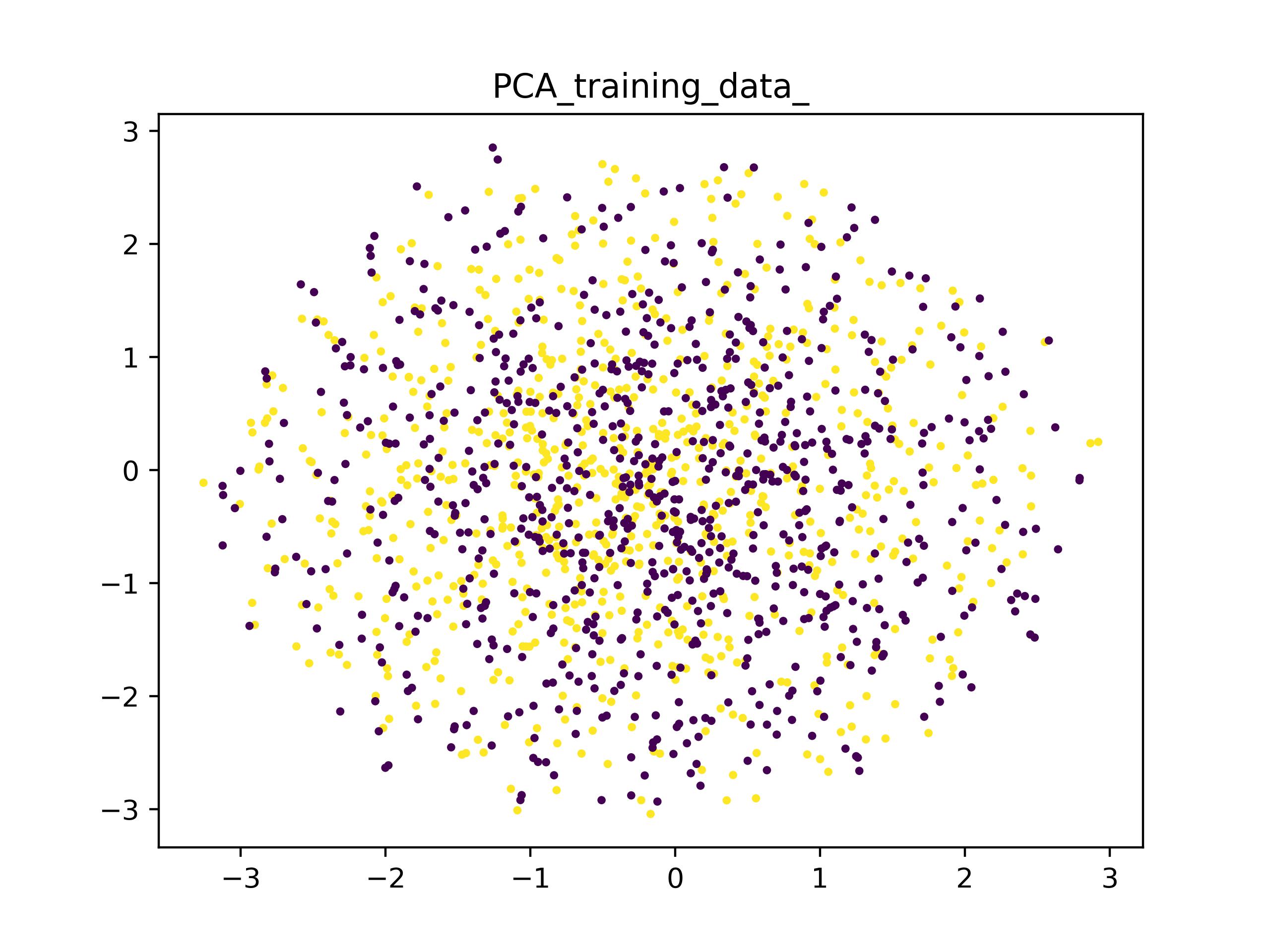} }}%
    \qquad
    \subfloat[\centering Bad choice of $\vec{\sigma}$]{{\includegraphics[width=5cm]{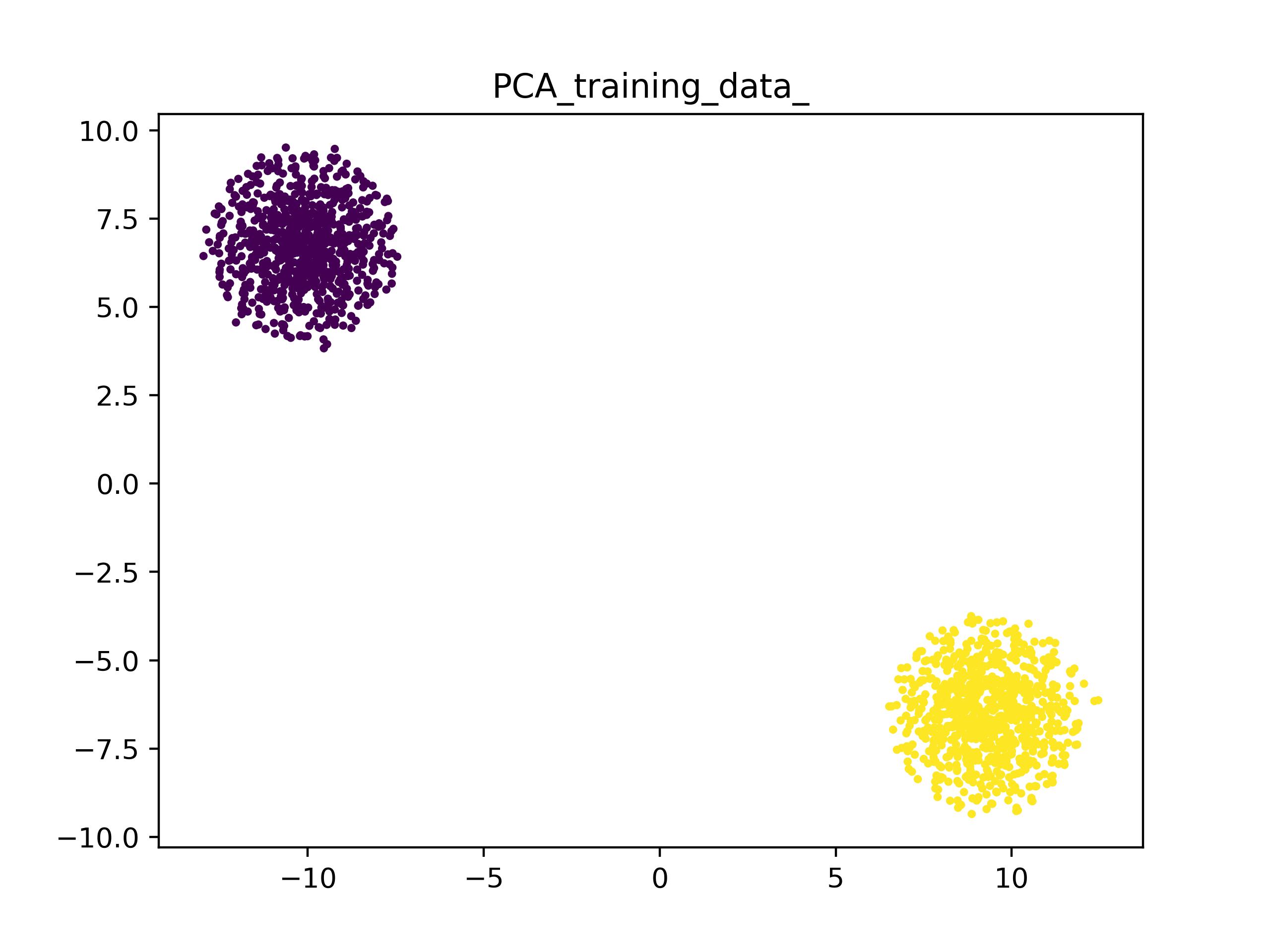} }}%
    \caption{Example of two synthetic datasets, a \textit{good one}, and a \textit{bad one}}
    \label{fig:synthetic_pca}%

\end{figure}

The requirements of this synthetic dataset are enumerated below along with the design decisions behind them:
\begin{enumerate}
    \item The dataset will contain objects from \textbf{two classes} as we want to perform a classification task in the final stage to assess the quality of our lower-latent space embeddings
    \item To perform \textbf{data integration}, the objects will be described by two modalities, where each modality is sampled from a different normal distribution
    \item The distributions used to generate the modality data must be \textbf{intractable for a Principal Component Analysis} i.e. there must be no clear separation of classes after appying PCA, such as in Figure \ref{fig:synthetic_pca} (a), because having the objects already clustered would defeat the purpose of the experiment
    \item To easily build edges between samples of same class (intra-class) and samples belonging to different classes (inter-class) in order to evaluate how various graph building algorithms reflect on the quality of the lower-space embeddings
\end{enumerate}

After considering these matters, we decided that for each class of objects we should sample points with features coming from two multi-Gaussian with high standard deviation, such the feature spaces would overlap with the others class feature space. Let $A$ and $B$ be my two classes of labels. Let $\alpha$ and $\beta$ my two modalities. Now, let's assume feature space for $\alpha$ has $n$ dimensions, and $\beta$'s feature space has m. Let's define $\vec{\mu_{\alpha}}$ and $\vec{\mu_{\beta}}$, with $\mu_{\alpha, i}$ and   $\mu_{\beta, i}$ are coming from a two uniform distribution with different parameters, that can be freely chosen. We will define $\vec{\mu}_{\alpha, A} = \vec{\mu}_{\alpha} - \theta_{\alpha}$, where $\theta_{\alpha}$ can be again chosen by as, and $\vec{\mu}_{\alpha, B} = \vec{\mu}_{\alpha} + \theta_{\alpha}$. For the second modality the process is similar, by replacing $\alpha$ with $\beta$. 
 
Now we sample $X_{\alpha, A}$ from $\mathcal{N}( \vec{\mu}_{\alpha, A}, \vec{\sigma})$, $X_{\beta, A}$ from $\mathcal{N}( \vec{\mu}_{\beta, A}, \vec{\sigma})$,  $X_{\alpha, B}$ from $\mathcal{N}( \vec{\mu}_{\alpha, B}, \vec{\sigma})$ and $X_{\beta, B}$ from $\mathcal{N}( \vec{\mu}_{\beta, B},\vec{\sigma}) )$ where $\vec{\sigma}$ is big enough to cause overlap between each modalities feature space. For example, in Figure (a) we have good choice of $\vec{\sigma}$, but in Figure (b) we cannot say the same thing anymore. 

\section{Synthetic Graph}
\begin{figure}[h!]
    \centering
    \includegraphics[width=15cm]{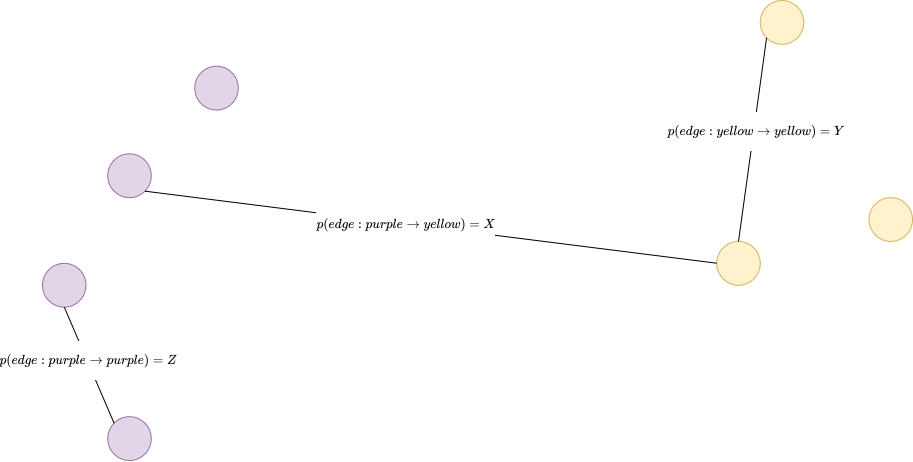} 
    \caption{Building a graph from the synthetic dataset}%
    \label{fig:synthetic-graph}%
\end{figure}

In order to build node relations between the data points coming from the Synthetic dataset, we decided to implement a statistical approach over the task. Let purple and yellow be the two possible labels. Our interest was to manipulate graph configurations in such a manner that we would be in control of the number of the edges between nodes with same label and with opposite label. And this is important for many reasons, we will latter explain, but for now the reader must just trust that the performance of the models applied over the dataset will be heavily influenced by this fact. As you can see in \ref{fig:synthetic-graph}, we will build edges between two purple nodes with probability $Z$, between purple and yellow with probability $X$, and between yellow nodes with probability $Y$. In the experiments and evaluation section we wanted to show of how model's performance can be influenced by different graphs structures. 

While, it might seem a counter-intuitive to build edges this way, it can be rationalized in the following way: if $p(edge: purple \to purple) = 1/2$ and  $p(edge: yellow \to yellow) = 1/2$ then the $p(edge: purple \to yellow) = 0$ (we would get two isolated graphs), or if $p(edge: purple \to purple) = 1/3$ and  $p(edge: yellow \to yellow) = 1/3$ then the $p(edge: purple \to yellow) = 1/3$. This is the probabilistic approach.

As we mentioned we will used a statistical approach which works the other way around. We will generate random samples of edges between purple-to-purple, purple-to-yellow, yellow-to-yellow nodes and this their number will be x, y , and y. In order to get back to the probabilistic approach, we need to compute $X = \frac{x}{x+y+z}$, $Y = \frac{y}{x+y+z}$, and $Z = \frac{z}{x+y+z}$. 
If we want to obtain $p(edge: purple \to purple) = 1/2$ and  $p(edge: yellow \to yellow) = 1/2$ then the $p(edge: purple \to yellow) = 0$ probabilities, we will just generate for example, $2000$ edges between purple-to-purple nodes and $2000$ edges between yellow-to-yellow nodes. 
Our decision to build graphs this way, will make further explained in the chapter Graph Neural Networks for Cancer Data Integration. 
\chapter{Deep Neural Networks}
This chapter provides a summary of the most prominent neural network architectures, their potent variants and applicability. We will first introduce the multi-layer perception as a knowledge base for the following models leveraged in this work: Variational Autoencoders (VAE), and Graph Neural Networks (GNN). 

In the experiments detailed further in the paper these models have been used in an unsupervised fashion: rather than carrying out regression or classification tasks, the goal is to generate lower-latent space representations of the unlabeled data points. The Variational Autoencoder is at the core of the implemented models, and the convolutional layers are deeply explained as they come up in the graph learning techniques based on order-invariant convolutions. Finally, we introduce GNNs and describe the main approaches to performing unsupervised learning to generate lower-space embeddings of data points: Deep Graph Infomax, which maximizes mutual local information, and Variational Graph Autoencoder, which builds up from the traditional Variational Autoencoder with the addition that the decoder reconstructs the adjacency matrix. 

\section{Multilayer Perceptron}
Neural networks are Machine Learning models composed of simple processing units that can compute linear or nonlinear transformations (based on the activation function used) on vector inputs; these processing units are called perceptrons. One perceptron receives a vector $\vec{x} \in \mathbf{R^n}$, on which it applies a linear combination by multiplying with a weight vector $\vec{w} \in \mathbf{R^n}$ and adding a bias value $b$. Afterwards, a nonlinear function $\sigma$ can be applied to the result to obtain the final value for an output, $y$. The are a multitude of activation functions which are chosen based on the learning task. 
$$y = \sigma\left(b + \sum_{i=1}^{n}w_{i}x_{i}\right)= \sigma\left(b + \vec{w}^T\vec{x}\right)$$

Multilayer perceptron neural networks are formed of layers of perceptron units which receive inputs from previous layers, and apply the linear combination and activation functions described above to return the output values. For each individual layer, we can compute the output values with the matrix form as follows:
\begin{equation}
    \vec{y} = \sigma \left(W\vec{x} + \vec{b}\right)
\end{equation}

For a two layered perceptron neural network the formula is similar. On top of the $\vec{y}$ obtained in (2.1), we apply another linear transformation by multiplying with the second layer's weight matrix, adding its bias value and, finally, computing the same or a different activation function. The new result is:

\begin{equation}
    \vec{y} = \sigma \left(W\vec{y}_{previous} + \vec{b}\right) = \sigma \left(W\sigma_{previous} \left(W_{previous}\vec{x} + \vec{b}_{previous}\right) + \vec{b}\right)
\end{equation}

 Networks with more than one intermediate layer are called \textbf{deep} neural networks. A natural question that comes with such networks is: "How many layers?". It is argued by Cybenko in \cite{cybenko1989approximation} that any bounded continous real function can be approximated with only one layer and a sigmoidal activation layer. However, if this were the truth this chapter would end here, which is not case. There is no perfect architecture and finding a neural network that works for a specific kind of data is an engineering task where a lot of experiments and searching needs to be carried out, in order to find what works better. 
\section{Autoencoders}
\label{sec:vae}

Generally, an autoencoder is a model that consists of two networks: the encoder, which constructs lower-latent space embeddings from the data points, and the decoder, which reconstructs the input data. The encoder function  $E(\dot)$ takes $\theta$ as parameter, and the decoder function $D(\dot)$ takes $\phi$ as parameter. The lower-space embedding is learned from $E_{\theta}(x)$ and the reconstructed input is $y = D_{\phi}(E_{\theta}(x))$. The two parameters are learned together on the reconstructed data through a loss function chosen upon the underlying data, named reconstruction loss, which is Binary Cross Entropy (BCE) for  categorical data, or Mean Square Error (MSE) for continuous data. 

\begin{equation}
L_{MSE} (\theta, \phi) = \frac{1}{n} \sum_{i=1}^{n} (x_{i} - D_{\phi}(E_{\theta}(x_{i}))^2
\end{equation}

Let m be the number of classes, then we have: 
\begin{equation}
    L_{BCE} (\theta, \phi) = \frac{1}{n} \sum_{i=1}^{n} \sum_{j}^m x_{ij} log(D_{\phi}(E_{\theta}(x_{i}))
\end{equation}

Many variants of autoencoders have been proposed to overcome the shortcomings of simple autoencoders: poor generalization, disentanglement, and modification to sequence input models. Among these models are the Denoising Autoencoder (DAE) \cite{vincent2008extracting}, which randomly filters-out some of the input features to learn the defining characteristics of the input object; Sparse Autoencoder (SAE) \cite{coates2011analysis} adds a regularization term to the loss function to restrict the lower-latent representations from being sparse (i.e. many zero-valued entries in the feature vectors). Finally, Variational Autoencoder (VAE) \cite{kingma2013auto} is presented in the next section.

\subsection{Variational Autoencoders}
\begin{figure}[h]
    \small
    \centering
    \includegraphics[scale=0.60]{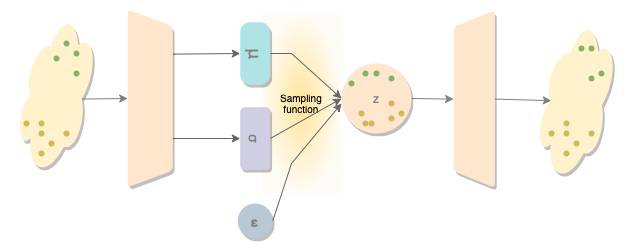}
    \caption{Generate low-dimensional representation of the input, regularize it in a Multi-Gaussian distribution, and attempt reconstruction of the original input}
    \label{fig:vae}
\end{figure}

Typically, VAE assumes that latent variables to be a centered isotropic multivariate Gaussian $p_{\phi}(z) = \mathbf{N}(z; 0, I)$, and $p_{\theta}(z|x)$ a multivariate Gaussian with parameters approximated by using a fully connected neural network. Since the true posterior $p_{\theta}(z|x)$ is untractable, we assume it takes the form of a Gaussian distribution with an approximately diagonal covariance. This allows the variational inference to approximate the true posterior, thus becoming an optimisation problem. In this case, the variational approximate posterior will also take a Gaussian with diagonal covariance structure:

$$q_{\phi}(z|x_{i}) = \mathbf{N}(z; \mu_{i}, \sigma_{i} I)$$

where $\mu$ and $\sigma$ will be outputs of the encoder. Since both $p_{\theta}(z)$ and $q_{\phi}(z|x_{i})$ are Gaussian, we can compute the discrepancy between the two:

\begin{equation}
 l_{i}(\theta, \phi) = - E_{q_{\phi}(z|x_{i})}[log p _{\theta}(x|Z)] + KL (q_{\phi}(z|x_{i})||p _{\theta}(z))
\end{equation}

with 

\begin{equation}
    KL(P(x)||Q(x)) =  P(x)log\left(\frac{P(x)}{Q(x)}\right)
\end{equation}

The first part of the loss function represents the reconstruction loss i.e. how different is the decoder output from the initial input, and the second part represents the reparametrisation loss. We used both Kullback-Leiber (KL) divergence (reconstruction loss which is used in the original paper describing Variational Autoencoders) and Maximum Mean Discrepancy (MMD) (which has been proved to give better by \cite{SimidjievskiEtAl2019}), which will be employed as an alternative to the KL divergence. While KL restricts the latent space embedding to reside within a centered isotropic multivariate Gaussian distribution, MDD is based on same principle, but uses the fact that two distributions are identical if, and only if, their moments are identical, with:

\begin{equation}
        MMD(P(x)||Q(x)) =  E_{p(x),p(x')}[k(x,x')] + E_{q(x),q(x')}[k(x,x')] -  2E_{q(x),q(x')}[k(x,x')]
\end{equation}

where $k(x,x')$ denotes the Gaussian kernel with $k(x,x') = e^{-\frac{||z-z'||^2}{2\sigma^2}}$. 
\begin{figure}[!h]
    \small
    \centering
    \includegraphics[scale=0.5]{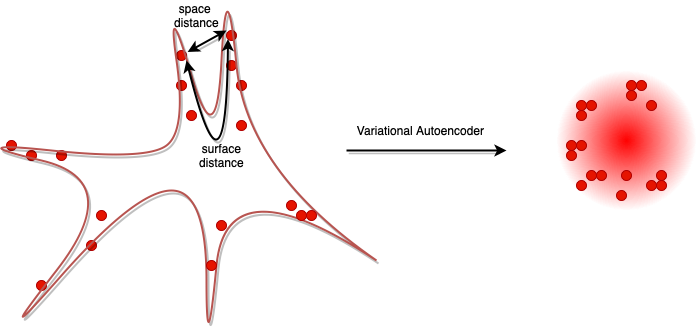}
    \caption{Although two points on the initial manifold might be close to each other space-wise, the distance on the actual surface might actually be larger, so a multivariate Gaussian representation of the data will 'flatten' the manifold's surface to better represent the similarity or disparity of the points. This is an intuitive picture.}
    \label{fig:vae}
\end{figure}
\section{Convolutional layers}
Datasets can take a plethora of shapes and forms as particular data sources are better described by different modalities, ranging from graphical and textual, to physiological signals and many other biomedical formats. Hence, the way we infer outcomes or define functions to describe the data must be adapted to the underlying characteristics of the dataset. For example, in most visual examples it is important to note that an area of neighbouring points present similar features, and leveraging this information helps in building a better performing model than just analysing all points in separation. This builds the case for convolutional layers that can summarise and learn functions for neighbourhoods of points, which is better suited for image-based datasets than multilayer perceptrons. Visual data generally has an $h \times w \times d$ shape where $h$ is the height, $w$ the weight, and $d$ the number of channels (e.g. colour images have three channels, RGB).

\begin{figure}[h]
    \small
    \centering
    \includegraphics[scale=0.5]{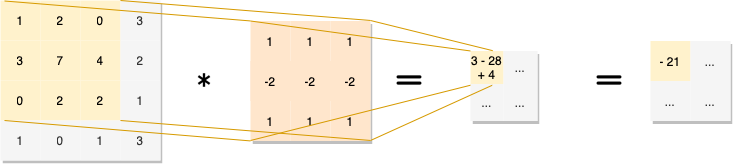}
    \caption{Convolutional layer applied on a single-channel data point}
    \label{fig:my_label}
\end{figure}

Let $X \in \mathbf{R}^{h \times w}$ and a kernel matrix $K \in \mathbf{R}^{n \times m}$ (ex. $m=n=3$). The new image $X'$ has the following formula:

$$X_{ab}' = \sum_{i=1}^{n} \sum_{j=1}^{m} K_{i j} X_{a+i-1,b+j-1}$$

\section{Graph Neural Network}
\label{sec:gnn}

This section presents Graph Neural Networks (or Graph Convolutional Networks, based on the source of the \cite{bojchevski2017deep}, \cite{bruna2013spectral}, \cite{defferrard2016convolutional}), which are learning techniques dealing with graph structured data that build intermediate feature representations, $\vec{x}'_i$, for each node \textit{i}. 

What must be noted is that all the architectures mentioned can be reformulated as an instance of message-passing neural networks \cite{kipf2016semi}.

\subsection{Graph Convolutional Layer}
Convolutional layers were introduced because most graph neural networks adopt \textit{the convolutional} element of it. Intuitively, when dealing with graph-like data we can say that nodes being neighbours with each other should be significant for the way we try to learn lower space embeddings. In our graph-shaped data, filters are applied to patches of neighbourhoods just as in CNN's example. These filters that we apply need to be order invariant because by taking the immediate neighbourhood of a node we cannot define a precise order of the neighbour nodes. 
\begin{figure}[h]
    \small
    \centering
    \includegraphics[scale=0.5]{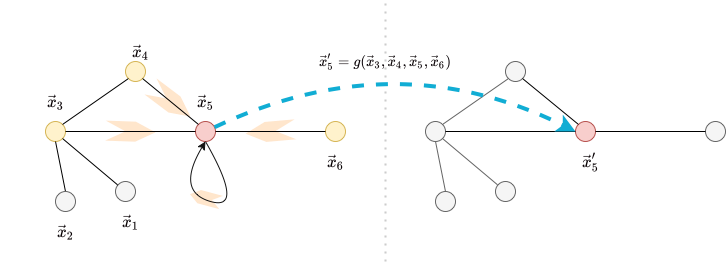}
    \caption{Convolutional layer applied on a single-node}
    \label{fig:my_label}
\end{figure}

Let $\mathcal{G} = (\mathcal{V}, \mathcal{E}, X)$ a graph where $\mathcal{V}$ is the vertex set, $\mathcal{E}$ is the set of edges, and X is the feature matrix, each row describing the features of vertex \textit{i} from $\mathcal{V}$. From $\mathcal{E}$ we can build $A$, the adjacency matrix, with following the rule: if $(i,j) \in \mathcal{E}$, then $A_{ij} = 1$ else $A_{ij}=0$. A simple way to \textit{aggregate} is to multiply X, the node feature matrix, with A. 
\begin{equation}
    X' = \sigma(AXW)
\end{equation}

where W is a parametrized learnable linear-transformation, shared by all nodes, and $\sigma$ is a non-linearity, an \textit{activation function}. A problem with this exact shape of the function is that after passing our inputs through it, we lose for each node it's own features, because $A_{ii} = 0$, as $(i,i) \notin \mathcal{E}$. A simple solution to this is to write $A' = A + I_n$ where $n = |\mathcal{V}|$. And now we have:

    \begin{equation}
        X' = \sigma(A'XW)
    \end{equation}

Because $A'$ may modify the scale of the output features, a normalisation is needed. So we define $D$ with $D'_{ii} = \sum_{j}A'_{ij}$, returning the degree matrix, 
    \begin{equation}
    \label{eq:2.4}
        X' = \sigma(D'^{-1}A'XW)
    \end{equation}

Node-wise, the same equation can be rewritten as below, which resembles \textit{mean-pooling} from CNNs:

    \begin{equation}
        \vec{x}'_{i} = \sigma(\sum_{j \in N_{i}} \frac{1}{|N_i|}W\vec{x}'_{j})
    \end{equation}

By using \textit{symmetric-normalisation} we get to the \textbf{GCN} update rule:

    \begin{equation}
    \label{eq:2.5}
        X' = \sigma(D'^{-\frac{1}{2}}A'D'^{-\frac{1}{2}}XW)
    \end{equation}

Which node-wise has following equation:
    \begin{equation}
        \vec{x}'_{i} = \sigma(\sum_{j \in N_{i}} \frac{1}{\sqrt{|N_i||N_j|}}W\vec{x}'_{j})
    \end{equation}
\subsection{Variational Graph Autoencoder}
Variational Graph Autoencoders and Graph Autoencoders were demonstrated by \cite{kipf2016variational} to learn meaningful latent representation on a link prediction task on popular citation network datasets such as Cora, Citeseer, and Pubmed. 
\begin{figure}[h]
    \small
    \centering
    \includegraphics[scale=0.45]{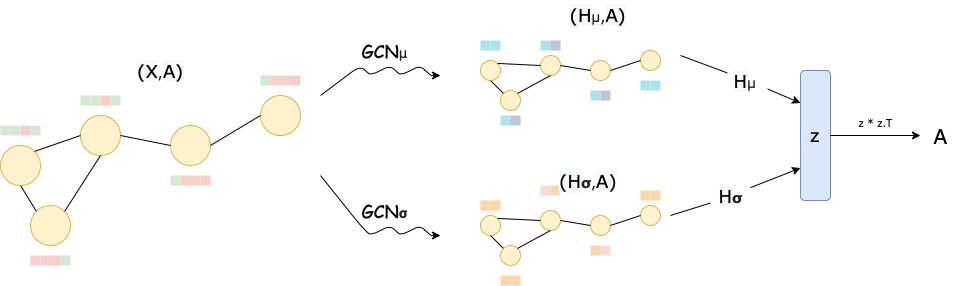}
    \caption{Convolutional layer applied on a single-channel point}
    \label{fig:my_label}
\end{figure}
Let $\mathcal{G} = (\mathcal{V}, \mathcal{\epsilon})$ an undirected and unweighted graph with $N = |\mathcal{V}|$ nodes. Let $\mathbf{A}$ be the adjacency matrix of $\mathcal{G}$. Node features are summarized in the vector $X \in \mathbf{R}^{N \times F}$, and $\mathbf{D}$ is the degree matrix. The authors further introduce the stochastic latent variables $\mathbf{z}_{i}$, summarized in $\mathbf{R}^{N \times F'}$

Similar to the Variational Autoencoder, a mean vector and logarithmic variation vector are produced, with two Encoder functions which can be composed by Graph Convolutional Layers, and then, by using a sampling function, the stochastic latent variables are reproduced. The loss function used for this learning task is exactly the same as the one used for a variational autoencoder, with the difference that for the reconstruction, the authors use inner dot product of the latent variables and compare the output with the input adjacency matrix. 

\begin{equation}
    \mathcal{L} = \mathbf{E}_{q(Z|X,Z)} [log_p(A|Z)] - KL(q(Z|X,A)||p(Z))
\end{equation}
As Z has $N \times F'$ dimension, we notice that $Z \cdot Z.T$ will have  $N \times N$ size, so it is possible to apply a loss function on $Z \cdot Z.T$ and adjacency matrix.

\subsection{Deep Graph Infomax}
Deep Graph Infomax was first described by Velickovic in \cite{velickovic2019deep}. The approach is based on maximizing local mutual information, and was inspired from \cite{hjelm2018learning} Deep Infomax.

For a generic graph-based unsupervised Machine Learning task setup, we will use the following notations. Let $X = \{\vec{x_{1}}, \vec{x_{2}}, ... \vec{x_{n}}\}$  be node feature set, where N is the number of nodes in our graph. Let $A\in \mathbf{R^{N \times N}}$ be the adjacency matrix with $A_{i,j}=1$ if and only if there is an edge between node i and j. The objective is to learn an \textit{encoder}, $\epsilon : \mathbf{R^{N\times F}} \times \mathbf{R^{N\times N} }\to \mathbf{R^{N\times F'}} $, such that $\epsilon (X, A) = H = \{\vec{h_{1}}, \vec{h_{2}}, ... \vec{h_{n}}\},$ The representations can be latter used for a classification task, and this will also represent a way we can evaluate the quality of our embeddings.

\begin{figure}[h]
    \small
    \centering
    \includegraphics[scale=0.4]{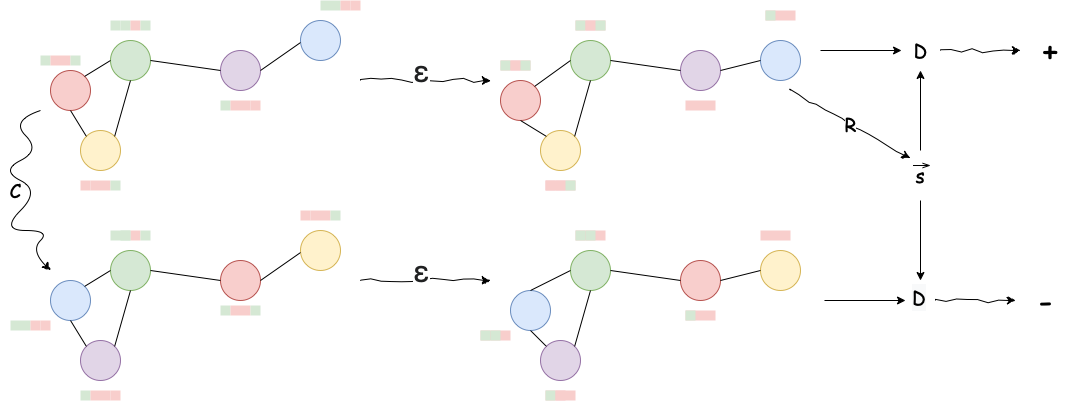}
    \caption{High-level overview of Deep Graph Infomax}
    \label{fig:my_label}
\end{figure}
In order to obtain graph-level summary vectors, $\vec{s}$, the authors leverage a readout function $R : \mathbf{R}^{N \times F} \to \mathbf{R}^{F}$ to summarise the obtained patch representation into a graph-level representation, $\vec{s} = R(\epsilon(X))$. For maximizing the local mutual information, $\mathit{D} : \mathbf{R}^{F} \times \mathbf{R}^{F} \to \mathbf{R}$, the \textit{discriminator} is deployed. $\mathit{D(\vec{h_{i}, \vec{s}})}$ should score higher if the patch representation is found in the summary.

The negative samples for D, are computed with a \textit{corruption} function $\mathit{C} : \mathbf{R}^{N\times F} \times \mathbf{R}^{N\times N} \to  \mathbf{R}^{M\times F} \times \mathbf{R}^{N\times N} $. The choice of the corruption function governs the specific kind of information that will be maximized. In my case, I have solely used a simple shuffling of the nodes features, letting the edges to stay in place. $(\tilde{X},\tilde{A})=\mathit{C}(X,A) = (X_{shuffled}, A)$ for this precise case.

The authors followed the original paper, which was not concerned with graph shaped like data \cite{hjelm2018learning}, and use a noise-constrastive type object with with a standard binary cross entropy loss between the samples from the joint and product of the marginals. 

\begin{equation}
    \mathit{L} = \frac{1}{N+M} \left( \sum_{i=1}^{N} \mathbf{E}_{(X,A)}\left[log \mathit{D}\left(\vec{h}_{i}, \vec{s}\right)\right]+ \sum_{j=1}^{M} \mathbf{E}_{(\tilde{X},\tilde{A})}\left[log\left(1 - \mathit{D}\left(\vec{h}_{i}, \vec{s}\right)\right] \right)\right)
\end{equation}

\chapter{Recreating Two State-of-the-Art Models}
\section{Description of Models}
The authors in ``Variational Autoencoders for Cancer Data Integration: Design Principles and Computational Practice" \cite{SimidjievskiEtAl2019} use several variational autoencoder architectures to integrate the METABRIC subsets containing CNA, mRNA and clinical data; the approaches in the paper are evaluated by combining pairs of modalities. We reproduce two models from this work, specifically CNC-VAE (Concatenation Variational Autoencoder) and H-VAE (Hierarchical Variational Autoencoder), both based on the VAE architecture. The difference between the two designs is where in the model the data integration is performed, with CNC-VAE concatenating the input features at the earliest stage, and H-VAE using hierarchical ordering, i.e. lower-latent space representations will be learned for each modality in part by a separate lower-level VAE, after which another higher-level VAE is applied on concatenation of the previously learned representations.

Generally, the integrative models obtain higher results than the raw data with CNC-VAE  obtaining accuracies as high as 82\% on PAM50 with an SVM classifier. X-VAE obtains good results on DR (77\%), and IC10 (85\%), while H-VAE obtains accuracies from DR (77\%) and PAM50 (82\%). 

\begin{figure}[!h]
    \small
    \centering
    \includegraphics[scale=0.40]{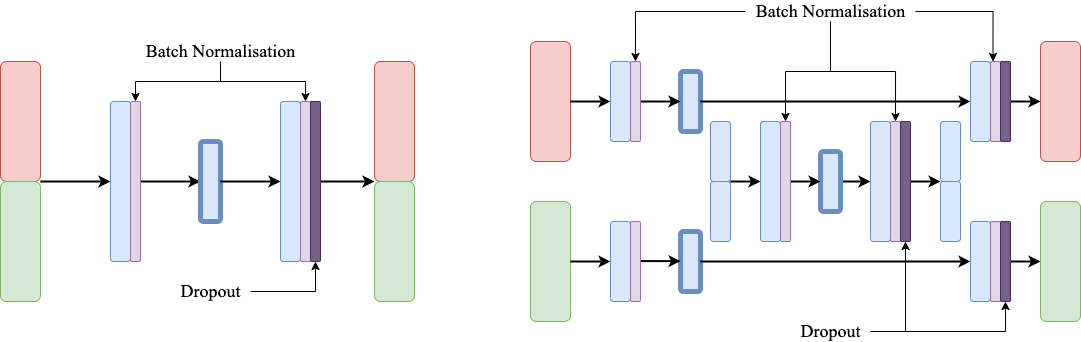}
    \caption{CNC-VAE and H-VAE. The two feature matrices are represented with red and green.}
    \label{fig:vaes}
\end{figure}

For \textbf{CNC-VAE}, the feature matrices of the inputs are concatenated and then fed to a Variational Autoencoder. The model is employed as a benchmark and as a proof-of-principle by the authors for learning a homogeneous representations from heterogeneous data sources. Even though early concatenation is prone to cause more noise, and modalities with lower-dimensional feature spaces will carry a weaker weight compared to the higher-latent feature vectors, this approach obtains competitive results (up to 84\%) with the other architectures. While the complexity of this simple architecture lies in the highly domain-specific data preprocessing, utilising a single objective function of combined heterogeneous inputs might not be ideal in other settings.

Unlike CNC-VAE, \textbf{H-VAE} learns lower-latent representations for all heterogeneous sources independently, and then concatenates the resulting homogeneous representations. The learning process takes place by training a lower-level Variational Autoencoder on each separate modality. After training these autoencoders, we concatenate the resulting lower-space embeddings for all modalities, and then train a higher-level Variational Autoencoder to learn the summarised embeddings of all intermediate representations. While for CNC-VAE we need to train only one network, for H-VAE we are going to train N + 1 neural networks, where N is the number of modalities. 

Both models use Batch Normalization (light violet) and Dropout(0.2) (dark violet) layers which are marked in Figure \ref{fig:vaes}. Dense layers use ELU activation function, with the exception of last dense layer which can also use according to case sigmoid activation(when the integration task is for CNA+Clin, categorical data). Where possible the reconstruction loss, is Binary Cross Entropy if data is categorical, and Mean Squared Error if data is continuous. For reparametrisation loss I have chosen MMD, because it gave significantly better results for the authors. Among the hyper-parameters we have the \textit{dense layer size} $ds$, the  \textit{latent layer size} $ds$, and $\beta$ the weight balancing between the reconstruction loss and the reparametrisation loss. 

$$\mathcal{L} = \mathcal{L}_{Reconstruction} + \beta \times \mathcal{L}_{Reparametrisation}$$

As an optimizer, all models use Adam with a learning rate of $0.001$
\section{Evaluation and results}
The environment in which I chose to reproduce CNC-VAE and H-VAE coming from \cite{SimidjievskiEtAl2019} is PyThorch \cite{paszke2019pytorch}, the originating one being TensorFlow with Keras \cite{abadi2016tensorflow}. I encountered a few challenges in the process of translating the models since some of the libraries in Tensorflow with Keras were outdated. Also, in the original code, the correct version of the H-VAE model was located in a different folder than the other models. 

The method of evaluation used is 5-fold cross validation. Each class of folds corresponding to a class of labels, being stratified, so making sure the distribution of labels over the folds is uniform.

\subsection{Hyper-parameter analysis}
For evaluating the quality of the reproduced models we carry out two experiments. The first one is performed to understand what hyper-parameter settings would be optional for each modality and each label, which is also the testing approach followed by the recreated paper \cite{SimidjievskiEtAl2019}. As we wanted to avoid repetition of results, the hyper-parameter search was done on Clin+CNA and DR (distance relapse) label. As noted by the authors, Naive Bayes classifier does not have any parameters, so it would be a good choice as the classifier used on top of our lower-space representations. 

\begin{figure}[!h]
    \small
    \centering
    \includegraphics[scale=0.46]{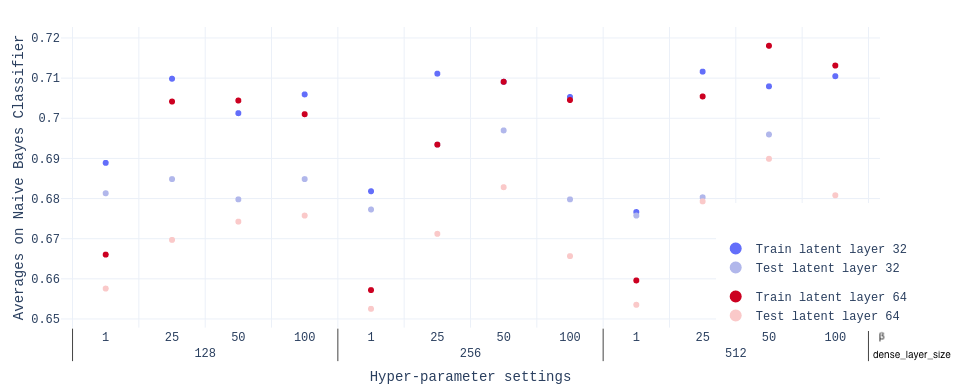}
    \caption{CNC-VAE performance with different hyper parameters settings.}
    \label{fig:cncvae_hp}
\end{figure}

\begin{figure}[!h]
    \small
    \centering
    \includegraphics[scale=0.46]{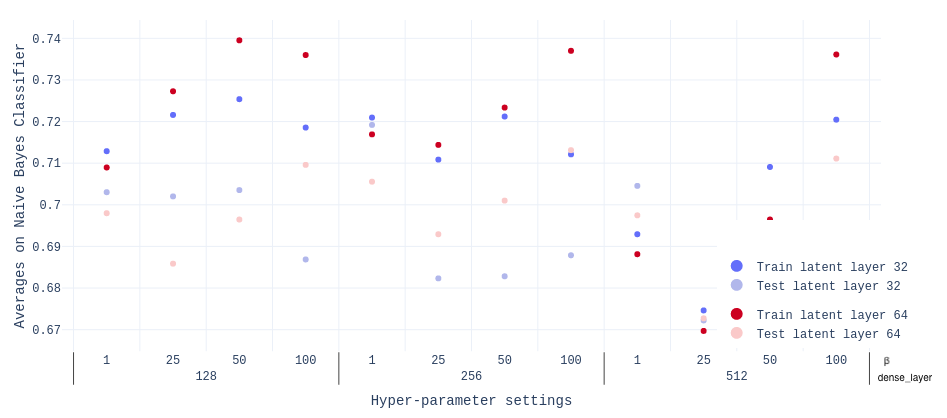}
    \caption{H-VAE performance with different hyper parameters settings.}
    \label{fig:hvae_hp}
\end{figure}

For my hyper-parameter setting I have picked $ds \in \{128, 256, 512\}$, $ls \in \{32, 64\}$, and $\beta \in \{1, 25, 50, 100\}$. The results can be seen in Figures \ref{fig:cncvae_hp} and \ref{fig:hvae_hp}.

\clearpage
\subsection{Best model assessment}
For the final experiment, we have picked fixed vales for \textit{ls}, \textit{ds}, $\beta$, and compared the accuracies of three classifiers: Naive Bayes, Support Vector Machine, and Random Forest, applied on the latent-lower space representations produced by H-VAE, and CNC-VAE. Because for $ls = 32$, $ds = 128$, $\beta = 25$ the results where generally good, I ran both my models with these parameters, and obtained the results in Table \ref{tab: Results reproduced models}. Generally, the results obtained in the original paper are better, but it must be noted that the aim of the authors was to fine-tune their models, while my goal was to show that I am able to reproduce models, and obtaining competitive results with the original ones. Another likely factor might is the difference in the learning time for H-VAE, which uses three different autoencoder networks. In our experiments, we allowed 150 epochs for each network.

\begin{table}[h]
\centering
\resizebox{\linewidth}{!}{%
\begin{tabular}{cccccccc}
\multicolumn{2}{c}{\multirow{2}{*}{}} & \multicolumn{3}{c}{CNC-VAE}             & \multicolumn{3}{c}{H-VAE}                \\
\multicolumn{2}{c}{}                  & CNA + mRNA  & Clin + mRNA & Clin + CNA  & CNA + mRNA  & Clin + mRNA & Clin + CNA   \\
\multirow{3}{*}{ER}    & NB           & 90          & 92          & 85          & 87          & 89          & 81           \\
                       & SVM          & \textbf{93} & \textbf{94} & 88          & \textbf{92} & \textbf{92} & 85           \\
                       & RF           & 88          & 90          & 83          & 87          & 88          & 80           \\
\multirow{3}{*}{DR}    & NB           & 66          & 69          & 70          & 67          & 68          & 70           \\
                       & SVM          & 68          & \textbf{71} & \textbf{70} & 62          & \textbf{69} & \textbf{72}  \\
                       & RF           & 67          & 69          & 56          & 67          & 69          & 69           \\
\multirow{3}{*}{PAM50} & NB           & 63          & 67          & 55          & 60          & 65          & 51           \\
                       & SVM          & \textbf{68} & \textbf{73} & 59          & \textbf{67} & \textbf{72} & 54           \\
                       & RF           & 62          & 67          & 54          & 57          & 58          & 47           \\
\multirow{3}{*}{IC}    & NB           & 68          & 74          & 59          & 66          & 62          & 53           \\
                       & SVM          & \textbf{75} & \textbf{79} & 63          & \textbf{73} & \textbf{73} & 56           \\
                       & RF           & 63          & 64          & 55          & 58          & 53          & 45          
\end{tabular}
}
\caption{Test results for classification with Naive Bayes (NB), Suport Vector Machine (SVM), and Random Forest (RF) of lower-latent representations produced by CNC-VAE and H-VAE, in percentages (\%). }
\label{tab: Results reproduced models}
\end{table}

Finally, we will discuss the contrast between the results we obtained and those in the original paper \cite{SimidjievskiEtAl2019}. Although we carried out our own hyper-parameter search on the same architectures, we arrived at a different setting that obtained better results on a small subset of modality and label class combinations but generally performed worse. Secondly, the implementation of our project was written in PyTorch, while the underlying Machine Learning framework leveraged in the reference paper was Tensorflow with Keras. Even though both frameworks overlap in supported functionalities overall, there are several methods that exist in Tensorflow but not in Pytorch, and custom implementations often have a minor impact on the model training. One of the main differences is training in batches, which comes out of the box with TensorFlow, but had to be manually implemented in PyTorch in our experiments. Another potential issue is the implementation of Maximum Discrepancy Loss function in PyTorch, because the variants found in other publications were different from the one written in TensorFlow, which was not directly transferable in PyTorch.

For fixed hyper-parameters, there is a total of $5(folds) \times 3 (modalities) \times 4 (labels) = 60$ models to train. For the hyper-parameters sets that the reference paper proposes and the two models, we would need to train $60 \times 108 = 6480$ different models. A model trains in cca. 2 minutes. That would be approximately 300 hours, which is 12 whole days to get best hyper-parameter settings for the two models. Thus, we only trained on a single fold to reduce the training time by five. For the best hyper-parameter setting we found, CNC-VAE clearly out-performs H-VAE.

\newtheorem{definition}{Definition}

\chapter{Graph Neural Networks for Cancer Data Integration}
\label{cha:graph neural networks for cancer data integration}

\begin{figure}[h!]
    \small
    \centering
    \includegraphics[scale=0.30]{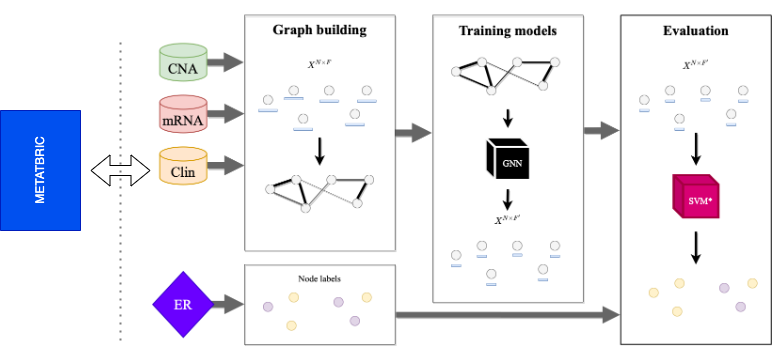}
\caption{This chapter is split in three modules: Graph Construction, Introducing novel Unsupervised Integrative Graph Neural Network, and Evaluate the quality of the lower latent space representations.}
    \label{fig:snf}

\end{figure}

This chapter presents the approaches used to generate lower-dimensional representations for multiple integrated modalities.  We will first introduce the graph construction algorithms along with the advantages and downsides, followed by the proposed integrative unsupervised models. Finally, we are evaluating the quality of the obtained lower-space representation. 

The data sets employed for graph construction are METABRIC and the synthetic dataset described earlier. The synthetic data will be used to discover the best settings for the proposed  Graph Neural Networks and to demonstrate the functionality of the proposed models. Furthermore, METABRIC will be leveraged in hyper-parameter fine-tuning for the graph construction modules thanks to the varied distribution of the four classes of labels (ER, DR, PAM, IC).

Finally, we will present best results obtained by the proposed models for each class of labels and then discuss conclusions in the final chapter. 

\section{Graph Construction Algorithms}
Graph Neural Networks require the input data to conform to a graph shape i.e. to have a matrix of features, $X$, containing all node information, and a matrix of adjacency, $A$. Since METABRIC does not store the relationship between patients , we need to define a module that builds graphs from feature matrix, X. The \textbf{quality} of the resulting graph will influence the final results - we will describe what graph quality is in a quantitative manner over the next sections, but to give the reader an initial intuition, the following question can be posed: ``Should nodes with the same or different labels be connected by edges?".

\subsection{From feature matrix to graph}
Assume a feature matrix $X \in \mathbf{R}^{N \times F}$, for which N is the number of objects and F is the number of features which define the space coordinates of the samples. To transition from a static data set to a graph, one needs to ``draw" edges between the objects, which will in turn be visualised as nodes. One naive but working solution is to link data points if they are ``close" to each other, where the metric describing closeness is the Euclidean distance. Assume a and b are the coordinates of two points A and B from X $$dist_{Euclidian}(A,B) = \sqrt{a^2 - b^2}$$

In specialised literature, the most popular ways to connect points in space that rely on Euclidean distance are:
\begin{itemize}
    \item Use a radius, $r$, as a threshold, and trace edges between nodes if the Euclidean distance between them is lower than $r$. 
    \item Use the K-Nearest Neighbours (KNN) method, which for a node A will return the $k$ nearest neighbours based on the Euclidean distance between nodes.
    \item Employ a combination of the two approaches presented above for different values of $k$ and $r$.
\end{itemize}

\begin{figure}[h]
    \small
    \centering
    \includegraphics[scale=0.7]{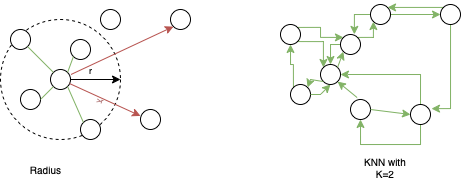}
    \caption{Graphic representation for choosing a threshold, $r$, for the radius method, or $k$ for KNN and the results of applying the two methods in separation}
    \label{fig:vae}
\end{figure}

Furthermore, to objectively assess the quality of the graphs presented in the next sections, we will introduce a metric cited by multiple sources in literature, namely homophily. Intuitively, it is employed to analyze the ratio of nodes connected by edges that have the same label.

\newpage
\subsection{Quantifying the quality of a graph}

\textbf{Homophily} is a metric that has been defined in many ways by different papers: edge homophily \cite{zhu2020beyond}, node homophily \cite{pei2020geom} and edge insensitive \cite{lim2021large}. In the context of this project we will refer mainly to edge homophily.
\begin{definition}
  (Edge Homophily) Given a $\mathcal{G} = \{\mathcal{V}, \mathcal{E}\}$ and a node label vector $y$, the edge homophily is defined as the ratio of edges that connect nodes with same label. Formally, it is defined as:
  
  \begin{equation}
      h (\mathcal{G}, { y_{i}, i \in \mathcal{V}}) = \frac{1}{|\mathcal{E}|} \sum_{(j,k) \in \mathcal{E}} \mathbf{1} (y_{j} = y_{k})
  \end{equation}

  where $|\mathcal{E}|$ is the number of edges in the graph and $ \mathbf{1} (A = B)$ returns $1$ if $A$ and $B$ have the same label, and $0$ if they do not.
\end{definition}

A graph is typically considered homophilous when $h(\cdot)$ is large (typically, $0.5<h(\cdot)<1$), given a suitable label context. Graphs with low edge homophily ratio are considered to be heterophilous. 

\begin{figure}[h]
    \small
    \centering
    \includegraphics[scale=0.50]{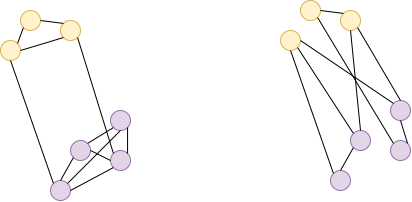}
    \caption{Left graph has a higher homophily than the right one, where the 'yellow' 'purple' colours represent the labels of the nodes}
    \label{fig:vae}
\end{figure}

The works in \cite{pmlr-v97-abu-el-haija19a}, \cite{chien2021adaptive}, \cite{cuzzocrea2018proceedings} argue that strong homophily is needed in order to achieve good performances. The next subsection presents an analysis of the homophily levels in different graph construction settings.

\subsection{Graphs build out of METABRIC modalities}
In this section, we analyse how different values for $r$ and $k$ influence the overall homophily levels for each modality in part and all label classes. As previously mentioned, homophily is a metric that measures how labels are distributed over neighbouring nodes (linked through an edge), hence, measure these levels is helpful because it creates an expectation for the lower-dimensional embeddings produced by the GNN. For example, if some related nodes belong to different labels, the lower-latent space representations will not be very accurate for those nodes.

The next sub-sections present the obtained homophily levels over the three modalities for the four classes of labels: ER, DR, PAM and IC. Each patient can be described in terms of ER+ or ER-, positive DR and negative DR, 5 sub-types of breast cancer - PAM, and 10 identified through research clusters - IC.

\subsubsection{Homophily levels for each class of labels on: Clinical data, and multi-omic data (mRNA, CNA), by using K Nearest Neighbours}

By looking at the tables below, we can learn the following aspects: in the KNN case, the homophily levels don't vary a a lot, in fact the remain at the same levels over an increase in the number of edges. In the case of the IC class of labels, notice that the results coming from CNA (40\%) and Clin (17\%) are very low.

\begin{table}[!h]
    \centering
    \small
    \begin{filecontents*}{csv/k/clin.txt}
K, Edges, {$}H_{ER}{$}, {$}H_{DR}{$},{$}H_{IC}{$}, {$}H_{PAM}{$}
4,7920,0.7742,0.6051,0.1674,0.3565
8,15840,0.7688,0.6063,0.1642,0.3504
12,23760,0.7670,0.6087,0.1645,0.3490
16,31680,0.7644,0.6073,0.1632,0.3484
24,47520,0.7630,0.6073,0.1615,0.3472
32,63360,0.7602,0.6057,0.1602,0.3437
48,95040,0.7533,0.6053,0.1581,0.3390
\end{filecontents*}
\csvautotabular[respect all]{csv/k/clin.txt}
    \caption{Homophily of graph build on \textbf{Clin}}
    \label{tab:clinK}

\end{table}    

\begin{table}[!h]
    \centering
    \small
    \begin{filecontents*}{csv/k/rnanp.txt}
K, Edges, H_{ER}, H_{DR}, H_{IC}, H_{PAM}
4,7920,0.8930,0.6118,0.5746,0.5632
8,15840,0.8859,0.6174,0.5491,0.5477
12,23760,0.8825,0.6149,0.5319,0.5393
16,31680,0.8773,0.6141,0.5200,0.5289
24,47520,0.8697,0.6151,0.5007,0.5166
32,63360,0.8643,0.6159,0.4853,0.5080
48,95040,0.8564,0.6180,0.4622,0.4946
\end{filecontents*}
\csvautotabular[respect all]{csv/k/rnanp.txt}
    \caption{Homophily of graph build on  \textbf{mRNA}}
        \label{tab:rnaK}

\end{table}

\begin{table}[!h]
    \centering
    \small
    \begin{filecontents*}{csv/k/cnanp.txt}
K, Edges, H_{ER}, H_{DR}, H_{IC}, H_{PAM}

4,7989,0.7867,0.5994,0.4508,0.3841
8,15898,0.7796,0.6059,0.4443,0.3841
12,23812,0.7725,0.6107,0.4373,0.3803
16,31728,0.7698,0.6066,0.4258,0.3804
24,47560,0.7636,0.6124,0.4133,0.3781
32,63392,0.7590,0.6147,0.4078,0.3753
48,95056,0.7505,0.6159,0.4004,0.3671
\end{filecontents*}
\csvautotabular[respect all]{csv/k/cnanp.txt}
    \caption{Homophily of graph build on \textbf{CNA}}
    \label{tab:cnaK}

\end{table}

\newpage
\subsubsection{Homophily levels for each class of labels on: Clinical data, and multi-omic data (mRNA, CNA), by using Radius R}

\begin{table}[!h]
    \centering
    \small
    \begin{filecontents*}{csv/r/clin.txt}
R, Edges, Iso. N, H_{ER}, H_{DR}, H_{IC}, H_{PAM}
1.2,420,1695,0.780,0.6380,0.1714,0.390
2.0,2998,1134,0.783,0.629,0.148,0.390
2.4,14756,573,0.779,0.630,0.152,0.367
2.52,35567,190,0.780,0.620,0.152,0.353
3.2,60041,5,0.728,0.598,0.152,0.318
3.6,62838,0,0.695,0.594,0.144,0.299
4,63322,0,0.673,0.596,0.139,0.284
5.0,63360,0,0.651,0.596,0.133,0.269
\end{filecontents*}
\csvautotabular[respect all]{csv/r/clin.txt}
    \caption{Homophily of graph build on \textbf{Clin}}
    \label{tab:clinR}
\end{table}    

\begin{table}[!h]
    \centering
    \small
    \begin{filecontents*}{csv/r/rnanp.txt}
R, Edges, Iso. N, H_{ER}, H_{DR}, H_{IC}, H_{PAM}
2.68,1680,1673,0.964,0.738,0.766,0.675
3.21,13784,1178,0.955,0.719,0.629,0.622
3.37,18548,1040,0.942,0.704,0.591,0.603
3.75,29694,701,0.909,0.677,0.504,0.540
4.28,45131,267,0.860,0.632,0.424,0.460
4.82,57380,44,0.821,0.604,0.405,0.432
5.36,62175,5,0.778,0.595,0.379,0.405
\end{filecontents*}
\csvautotabular[respect all]{csv/r/rnanp.txt}
    \caption{Homophily of graph build on  \textbf{mRNA}}
    \label{tab:rnaR}
\end{table}

\begin{table}[!h]
    \centering
    \small
    \begin{filecontents*}{csv/r/cnanp.txt}
R, Edges, Iso.N, H_{ER}, H_{DR}, H_{IC}, H_{PAM}
2.21,11181,1480,0.751,0.695,0.670,0.435
2.95,16673,1252,0.782,0.665,0.588,0.425
4.06,27634,907,0.800,0.634,0.431,0.408
4.65,33462,727,0.796,0.628,0.372,0.388
5.17,39124,525,0.794,0.614,0.321,0.371
5.91,48468,263,0.777,0.593,0.252,0.333
6.65,57666,73,0.751,0.585,0.205,0.296
\end{filecontents*}
\csvautotabular[respect all]{csv/r/cnanp.txt}
    \caption{Homophily of graph build on \textbf{CNA}}
    \label{tab:cnaR}

\end{table}

By looking at the tables we can learn following aspects. In the Radius case homophily levels vary a lot, sometimes even 40\% (in the mRNA $H_{IC}$ case). This means that or choice of R matters a lot. Another aspect that can be noticed is that for \textit{good} levels (above 70\%) of homophily, most of the time that graph conformation will have lots of isolated nodes, fact which can be disastrous for graph neural networks.

Now, with some intuition build on how graphs would behave like, we will reveal to the reader that a combination of the two methods will be used in order to get the most out of both. By using KNN we will ensure that no nodes are isolated, and by using the radius method we will ensure that the clusters of nodes close in space will be related through edges regardless of their number (limitation of the KNN).

The next section will present the proposed 4 models that attempt integration on graph structured data. While the fist two: CNC-VGAE and CNC-DGI attept early integration (by direct concatenation of the featuers), 2G-DGI and Hetero-DGI will attempt mixed integration, by concatenating their lower latent features in the middle of the learning phase. 
\section{Graph Neural Network Models For Data Integration}
We researched unsupervised learning approaches and models that could aid in the integration of two feature matrices that describe the same items in different ways for a data integration task. A first strategy would be to concatenate the two feature matrices and construct a graph on top of that, then apply either a Variational Graph Autoencoder or a Deep Graph Infomax. A second technique is to integrate two graphs on which will apply GCN layers, and then use a Dense Layer to 'filter' the two concatenated feature matrices during the integration phase. A third option is to create a hetero graph and concatenate the upper layer and bottom layer features at some point. Both the second and third method will imply at some point the use of Deep Graph Infomax.

We have selected the Deep Graph Infomax architecture type to integrate two graphs or a hetero graph. This is because for readout and discrimination the only inputs are latent feature vectors, with no adjacency matrix. The adjacency matrix can be very tricky to work with, and we will give two scenarios to prove our point. It is necessary to know that when applying a GCN layer over data points both their feature matrix and their adjacency matrix is needed.
\begin{itemize}
    \item  Consider applying two GCN layers to two graphs in order to integrate them. Then we wish to concatenate the resulted feature matrices and apply another GCN layer. A question is which adjacency matrix should we keep, the one from the first graph or the one from the second graph? Obviously the two graphs have different adjacency matrices.

    \item Another problem specific to VGAE is that, even if we can get lower-latent variables to incorporate the information from two graphs, when reconstructing, if we only use inner product we can only build one adjacency matrix, since
    the inner product of the lower latent-variables has no parameters to train. Alternatively, two layers could be added before generating the two adjacency matrices, and then rebuilt with the inner product. However, in the original paper the adjacency matrix is built directly from the latent space with a function that doesn't have any parameters.
\end{itemize}
\subsection{Notation and background knowledge}

Let $X_1^{N \times F_1}$ and $X_2^{N \times F_2}$ be the two modalities we want to integrate. Let $\mathcal{G} : R^{N \times F_i} \to  R^{N \times F_i} \times R^{|\mathcal{E}_i| \times 2} \times R^{|\mathcal{E}_i|}$, where $\mathcal{G}$ will return the feature matrix together with the edges set $\mathcal{E}_i$, and a list of their attributes.

Each architecture in part, will have a component $Encoder$, this will return the latent lower space representation and will have the following shape, with $ls$ the dimension of the lower latent space:

A graph encoder in this case will represent a function $$Encoder : \mathbf{R}^{N \times F} \times \mathbf{R}^{|\mathcal{E}| \times 2} \times R^{|\mathcal{E}|} \to \mathbf{R}^{N \times ls}$$
or  
$$Encoder : 2\times \mathbf{R}^{N \times F_i} \times \mathbf{R}^{|\mathcal{E}_i| \times 2} \times R^{|\mathcal{E}_i|} \to \mathbf{R}^{N \times ls}$$
for the Two Graphs Deep Infomax, which will be present in the next sections.

The encoder will be an ensemble of graph convolutional layers, plus some special integrative layer which will describe depending to case. Even thought there are a lot of current choices for the graph convolutional layers: ChebConv \cite{defferrard2016convolutional}, SAGEConv \cite{hamilton2017inductive}, GATConv \cite{velivckovic2017graph}, GCNConv, etc, because it permits for edge attributes, GCNConv has been chosen. The number of layers will represent a parameter, which will be studied in the upcoming sections by varying their number from 1-3, literature mentioning that 3 layers are usually ideal for encoding task \cite{velickovic2019deep}.

For an edge with size $s$ it's value of the feature will be $e^{-s}$. This is a good way to write edge features because edges that unite close neighbour points will have values between $1$ and $0.5$ and far apart nodes which will be united will have values bellow $0.1$. This mapping for edges features is also used by Similarity Fusion Network's paper \cite{wang2014similarity}.
\begin{figure}[!h]
    \small
    \centering
    \includegraphics[scale=0.5]{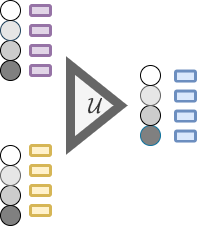}
    \caption{$\mathbf{\mathcal{U}}$: Special integration layer}
    \label{fig:special integration}
\end{figure}

The \textit{special integration layer} from \ref{fig:special integration} will be defined for each model in part (2G-DGI and Hetero-DGI), and will work \textbf{differently} for each of the two mentioned models.

In order to simplify notation, we will not mention \textit{edge attributes} in the following subsections, but they will be used in thetraining of the models.
\subsection{Concatenation Of Features: CNC-DGI and CNC-VGAE}
This method of integrating the two datasets is pretty straight forward. Having two feature matrices $X_{1} \in \mathbf{R}^{N \times F_{2}}$ and $X_{2} \in \mathbf{R}^{N \times F_{2}}$, the concatenation of them both would result in a matrix $X_{12} \in \mathbf{R}^{N \times (F_{1} + F_{2})}$. On top of this we apply our building graph method, and then we take our graph through the unsupervised method for getting lower space embeddings of our choice. 

\begin{figure}[!h]
    \small
    \centering
    \includegraphics[scale=0.4]{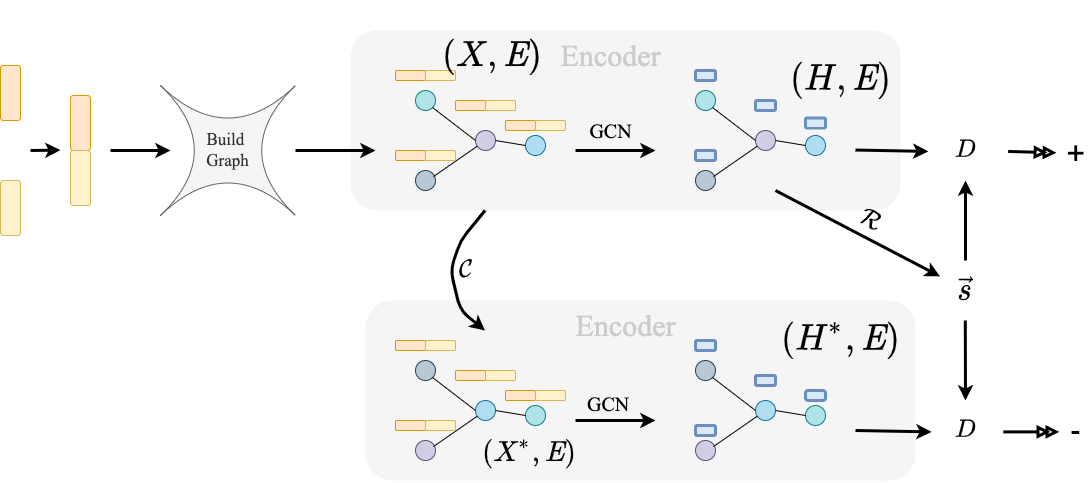}
    \caption{\textbf{CNC-DGI}: Apply Variational Autoencoder on top of the graph built on concatenated inputs.}
    \label{fig:cncdgi}
\end{figure}
\begin{figure}[!h]
    \small
    \centering
    \includegraphics[scale=0.43]{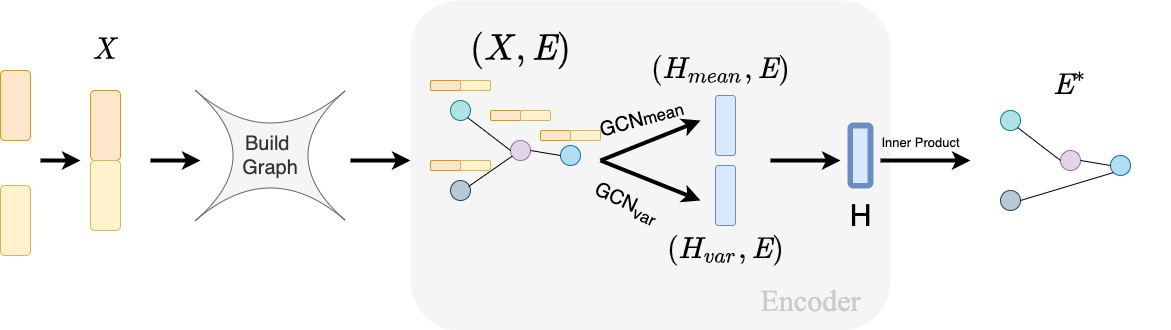}
    \caption{\textbf{CNC-VGAE}: Apply Variational Autoencoder on top of the graph built on concatenated inputs}
    \label{fig:cncvae}
\end{figure}

After training, the $\mathit{Encoder}$ from both will return lower-latent space embeddings with shape $\mathbf{R}^{N \times \mathit{ls}}$ where $\mathit{ls}$ is the dimension of the latent space. For \ref{fig:cncdgi} and \ref{fig:cncvae} the encoder will return $H^{N \times \mathit{ls}}$.

\subsection{Two Graphs: 2G-DGI}
Take $X_{1} \in \mathbf{R}^{N \times F_{1}}$ and $X_{2} \in \mathbf{R}^{N \times F_{2}}$ and build two graphs, $\mathcal{G}_{1} = ( X_{1}^{N \times F_{1}}, \mathit{E}_{1})$ and $\mathcal{G}_{2} = (X_{2}^{N \times F_{2}}, \mathit{E}_{2})$. The $GCN_{1} : R^{N \times F_1} \to R^{N \times \mathit{ls}} $ and $GCN_{2} :  R^{N \times F_2} \to R^{N \times \mathit{ls}} $ are different for $\mathcal{G}_{1}$ and for $\mathcal{G}_{2}$ because the nodes have different feature size.

Here the encoder, $\mathit{Encoder} : \mathbf{R}^{N \times F_{1}} \times \mathbf{R}^{N \times F_{2}} \to \mathbf{R}^{N \times \mathit{ls}}$ and $$\mathit{Encoder}(X_{1}, \mathit{E}_{1}, X_{2}, \mathit{E}_{2}) = \mathcal{U}(GCN_{1}(X_{1}, \mathit{E}_{1}), GCN_{2}(X_{2}, \mathit{E}_{2}))$$ where $\mathcal{U}$ can have for example the following shapes:
\begin{equation*}
   \mathcal{U}_{Dense} (GCN_{1}(X_{1}, \mathit{E}_{1}), GCN_{2}(X_{2}, \mathit{E}_{2}))=
   \end{equation*}
\begin{equation*}
   = Dense^{(2 \times \mathit{ls} \to \mathit{ls})}((GCN_{1}(X_{1}, \mathit{E}_{1}) ||  GCN_{2}(X_{2}, \mathit{E}_{2}))) =
\end{equation*}
\begin{equation}
  = Dense^{(2 \times \mathit{ls} \to \mathit{ls})}(H_1||H_2) = H
\end{equation}

\begin{equation}
    \mathcal{U}_{avg} (GCN_{1}(X_{1}, \mathit{E}_{1}), GCN_{2}(X_{2}, \mathit{E}_{2})) = \frac{GCN_{1}(X_{1}, \mathit{E}_{1}) + GCN_{2}(X_{2}, \mathit{E}_{2})}{2} = \frac{H_1 + H_2}{2} = H
\end{equation}
One observation worth making is that while $\mathcal{U}_{Dense}$ has parameters, $\mathcal{U}_{Avg}$ does not. We propose those two different \textit{special integration layers} because we can learn what works better in the DGI context. 

\begin{figure}[h]
    \small
    \centering
    \includegraphics[scale=0.27]{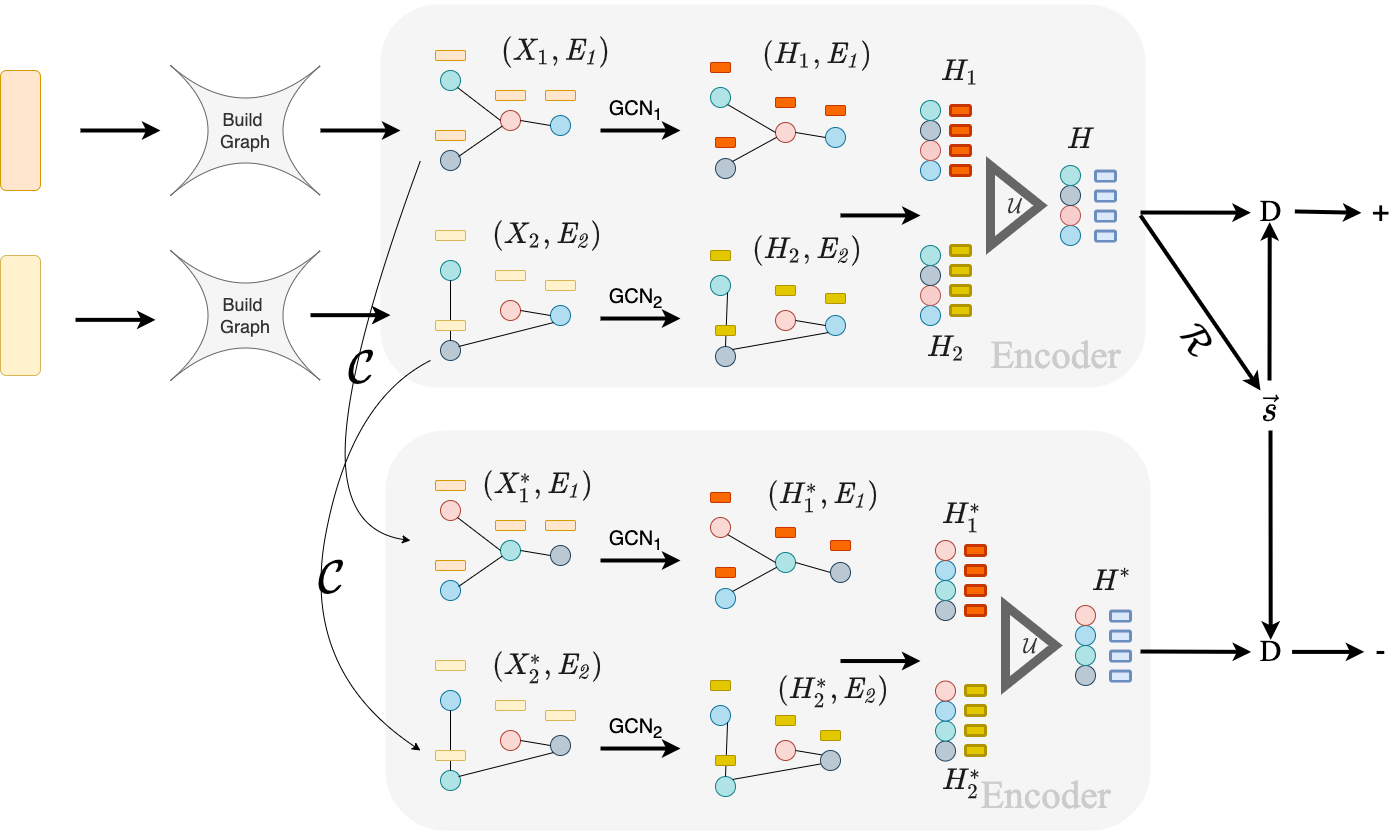}
    \caption{\textbf{2G-DGI}: Two graphs integration}
    \label{fig:2GDGI}
\end{figure}
\clearpage
\subsection{Heterogeneous Graph: Hetero-DGI}
In \cite{wang2019heterogeneous} heterogeneous graphs have the following definition:

\begin{definition}
  A \textbf{heterogeneous graph} denoted as $\mathcal{G} = ( \mathcal{V}, \mathcal{E})$ , consists of an object set $\mathcal{V}$ and a link set $\mathcal{E}$. A heterogeneous graph is also associated with a node type function $\theta : \mathcal{V} \to \mathcal{A}$ and a link type mapping function $\omega : \mathcal{E} \to \mathcal{R}$. $\mathcal{A}$ and $\mathcal{R}$ denote two sets of predefined object types and link types, where $|\mathcal{A}| + |\mathcal{R}| > 2$.
\end{definition}
Take $X_{1} \in \mathbf{R}^{N \times F_{1}}$ and $X_{2} \in \mathbf{R}^{N \times F_{2}}$ and build two graphs, $\mathcal{G}_{1} = ( X_{1}^{N \times F_{1}}, \mathit{E}_{1})$ and $\mathcal{G}_{2} = (X_{2}^{N \times F_{2}}, \mathit{E}_{2})$.  In order to get our heterogeneous graph we will add edges between the nodes that describe same objects, and we will say these edges belong to $\mathcal{E}_{3}$. Now, the two node types are defined by the graph the node is originating from. For edges, $\mathit{e}$ is of type $i$ if it belongs to $\mathcal{E}_{i}$. 

In here $\mathit{Encoder} : \mathbf{R}^{2N \times F} \to \mathbf{R}^{N \times \mathit{ls}}$
\begin{equation}
\mathit{Encoder}(X, \mathit{E}) = \mathcal{U}(GCN(X, \mathit{E}))
\end{equation}
In here, $\mathit{U}$ must do more than just concatenate. We have $\mathit{U} : \mathbf{R}^{2N \times ls} \to \mathbf{R}^{N \times ls}$. So we must have define a  split function $\mathit{S} : \mathbf{R}^{2N \times ls} \to \mathbf{R}^{N \times ls} , \mathbf{R}^{N \times ls}$, split the feature matrix on the nodes axes. Next we will define $\mathcal{U}_{Avg}$ and $\mathcal{U}_{Dense}$:
\begin{equation}
\mathcal{U}_{Avg} (GCN(X,\mathit{E})) = \frac{{\sum(S(GCN(X,\mathit{E})))}}{2} = \frac{H_1+H_2}{2} = H
\end{equation}
\begin{equation*}
\mathcal{U}_{Dense} (GCN(X,\mathit{E})) = Dense^{(2 \times \mathit{ls} \to \mathit{ls})}(||(S(GCN(X,\mathit{E})))) = 
\end{equation*}
\begin{equation}
 = Dense^{(2 \times \mathit{ls} \to \mathit{ls})}(H_1||H_2) = H
\end{equation}

Just as in previous case, we are taking two $\mathcal{U}$: a parametric one, and a fixed one. In the evaluation section 
\begin{figure}[!h]
    \small
    \centering
    \includegraphics[scale=0.35]{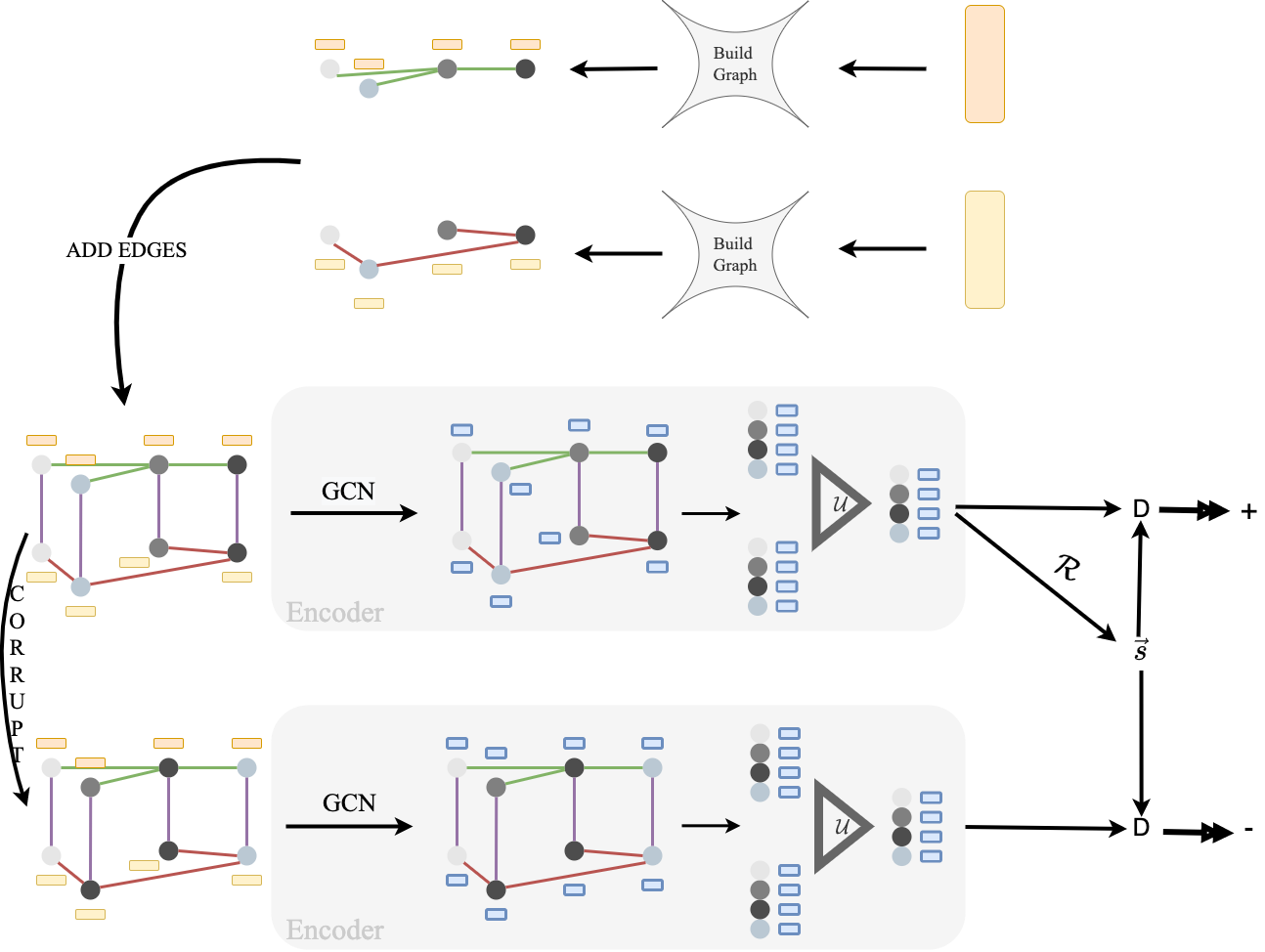}
    \caption{Build graph on concatenated features}
    \label{fig:vae}
\end{figure}
\clearpage
\section{Evaluation and results}

In order to evaluate the quality of the proposed models, we have decided to proceed with two testing methods. Evaluation of the models on  METABRIC dataset has one big problem which is that for one hyper-parameter setting 60 models need to be trained, in order to correctly evaluate the models performance. We thought that a synthetic dataset which needs to train only one model in order to asses the quality of a hyper-parameter setting  would be more fitted, since it has only one label class would be helpful, and only one combination of modalities which can be integrated.  

\begin{figure}[!h]
    \small
    \centering
    \includegraphics[scale=0.60]{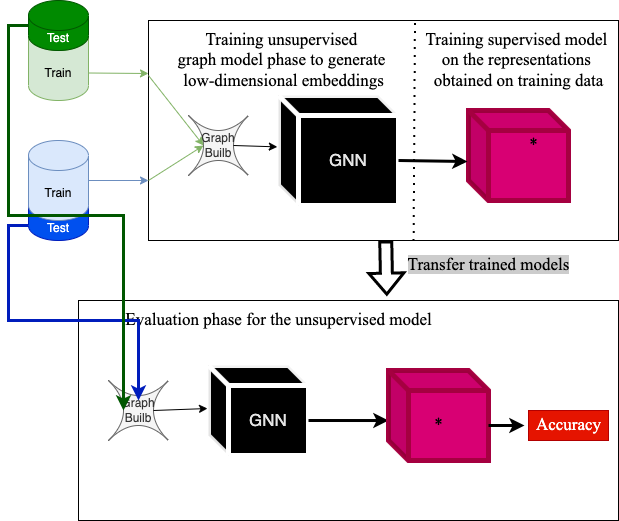}
    \caption{Evaluation pipeline for this project}
    \label{fig:2GraphsDGI_AVG}
\end{figure}

The supervised models in use will be: Naive Bayes \cite{murphy2006naive}, Support Vector Machine \cite{noble2006support} and Random Forest \cite{qi2012random}.
\section{Evaluation on Synthetic-Data}
In order to test these models on synthetic data, we have split each modality of the synthetic dataset in Training and Testing, with 75\% of the samples for training and 25\% of the samples for testing. While, a five-fold cross validation, would have been more suitable, the number of models we would have tested with different hyper-parameter settings would have been $\times 5$, because we need to re-train each model when the fold changes. The next subsections, will present the evaluation of CNC-DGI, CNC-VGAE, 2G-DGI and Hetero-DGI for various parameters. 
\subsubsection{CNC-DGI}
\begin{figure}[!h]
    \small
    \centering
    \includegraphics[scale=0.33]{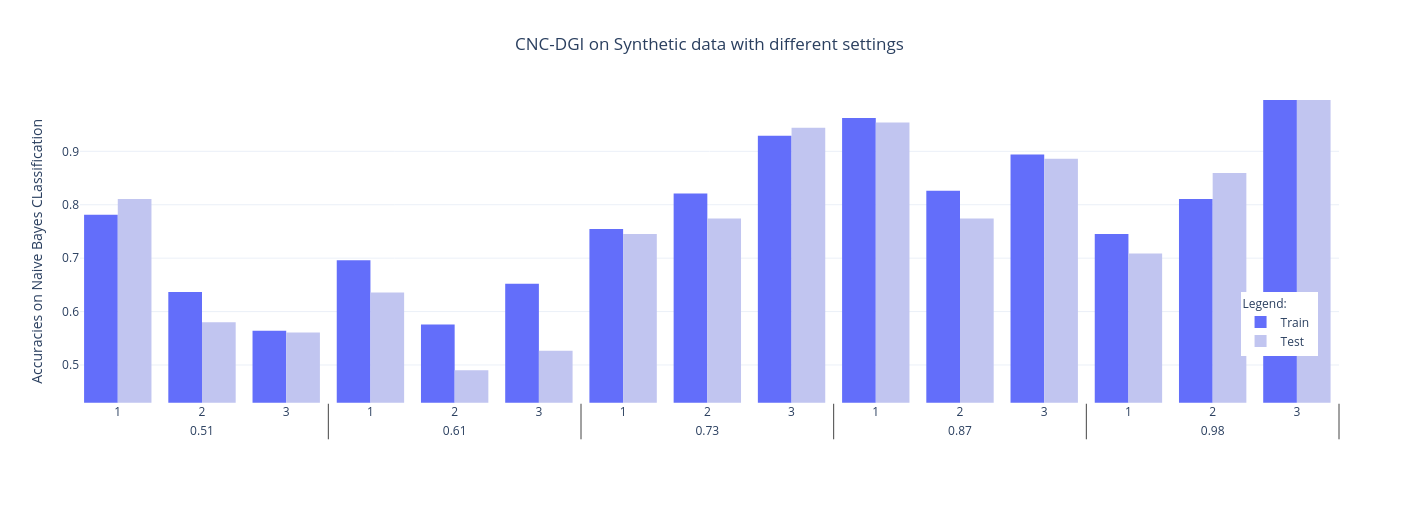}
    \caption{Accuracies of lower-latent representation obtained from CNC-DGI }
    \label{fig:2GraphsDGI_AVG}
\end{figure}
For CNC-DGI the parameters chose are the depth of the Encoder, i.e. how many convolution layers there were going to be used $conv_{nr} = \{1,2,3\}$, in the context of five homophily levels $homophily_{levels} =\{0.51, 0.61, 0.73, 0.87, 0.98\}$. 

\subsubsection{CNC-VGAE}
\begin{figure}[!h]
    \small
    \centering
    \includegraphics[scale=0.30]{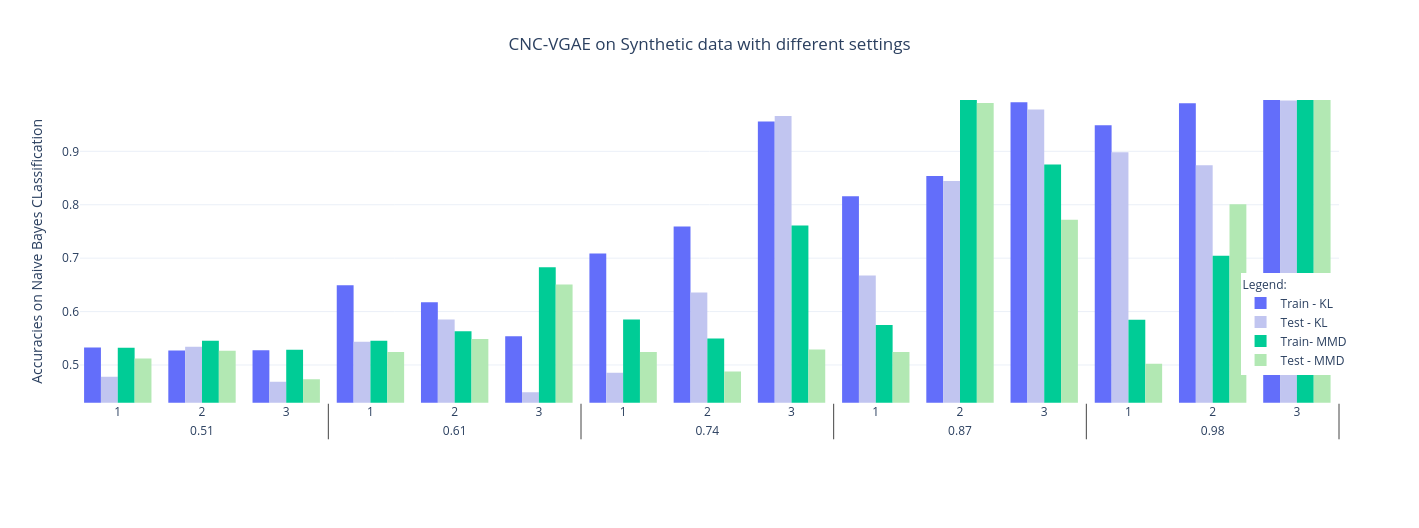}
    \caption{Accuracies of lower-latent representation obtained from CNC-VGAE}
    \label{fig:2GraphsDGI_AVG}
\end{figure}
For CNC-VGAE the parameters chose are the depth of the Encoder $conv_{nr} = \{1,2,3\}$, and the reparametrisation loss function, which could be either MMD or KL diveregence loss, in the context of five homophily levels.
From this diagram we can see that KL loss is more persistent as homophily levels increase, and as the number of layers increase. The best configuration for CNC-VAE can be with 3 GCN layers and using KL reparametrisation loss, even though \cite{SimidjievskiEtAl2019} use MMD for reparametrisation loss. at the same time METABRIC homophily levels will be quite small maybe bellow 50\% and from this diagram we can see that MDD gives better accuracies than KL on smaller homophily levels. 
\subsubsection{2G-DGI}
For the 2G-DGI, the parameters chose where the number of convolution used from 1 to 3, and the concatenation layer shape (either dense layer or average), in the context of different homophily levels in $\{0.51, 0.67, 0.74, 0.97\}$
\begin{figure}[!h]
    \small
    \centering
    \includegraphics[scale=0.35]{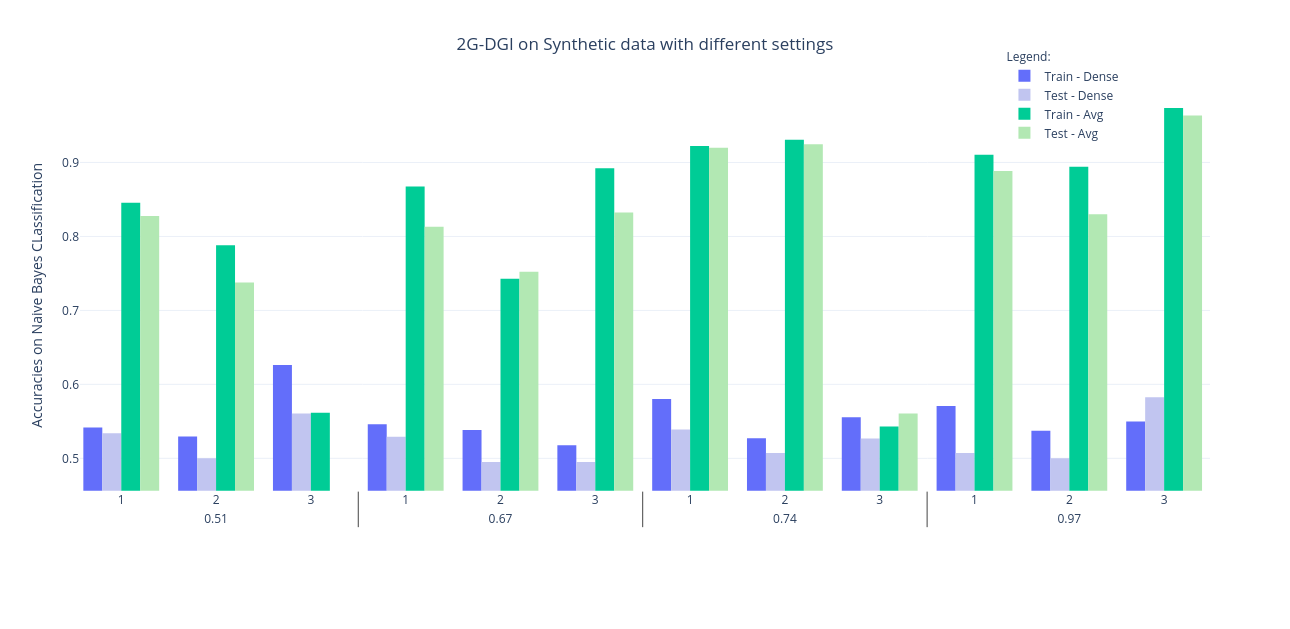}
    \caption{Accuracies of lower-latent representation obtained from 2G-DGI}
    \label{fig:2GraphsDGI_AVG}
\end{figure}

One can see that, generally the results obtained with two GCN layers and with an average filter are better than the results obtained with a dense layer on other depths on the encoder, over various levels of homophily.

\subsubsection{HeteroDGI}
\begin{figure}[!h]
    \small
    \centering
    \includegraphics[scale=0.34]{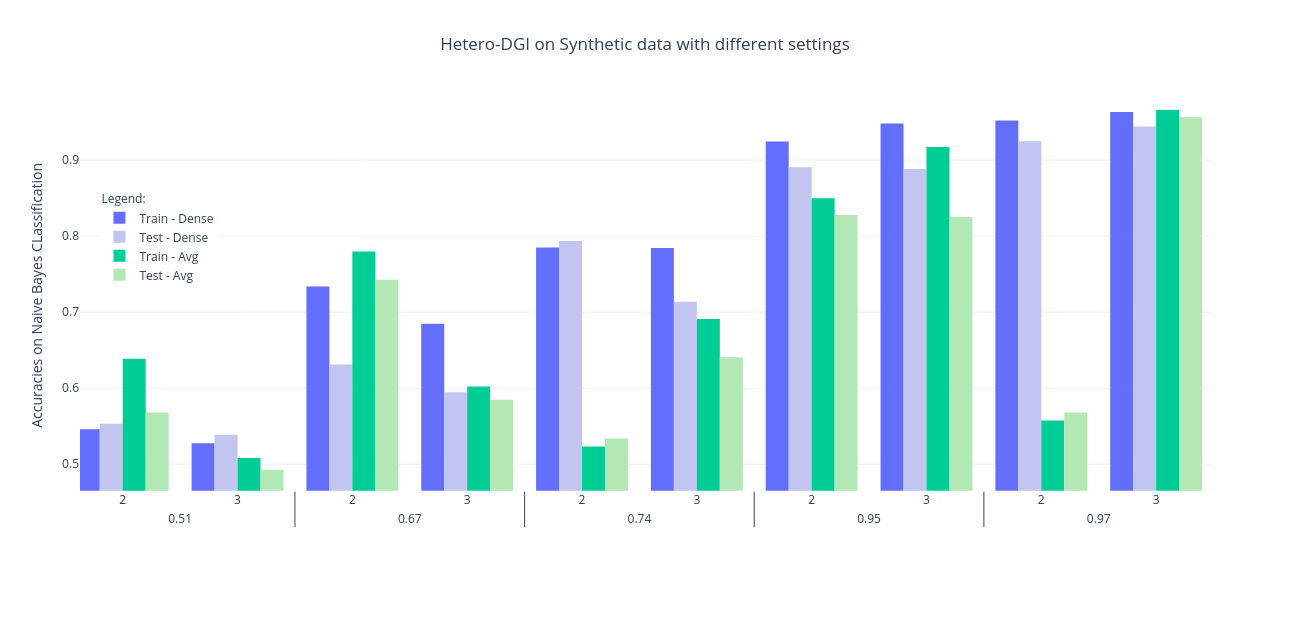}
    \caption{Accuracies of lower-latent representation obtained from Hetero-DGI}
    \label{fig:2GraphsDGI_DENSE}
\end{figure}

For the Hetero-DGI the evaluation setting is similar with 2G-DGI with the difference that encoders with one convolutional layers have been ignored. From the results one can learn that generally encoders with two convolutional layers work and with an dense layer give better results than all the other settings over various levels of homophily. 

\subsubsection{Conclusions}
From these diagrams, one can learn that the homophily levels will greatly influence the quality of the lower-latent space representations produced by the proposed models. This means that when looking for best hyper-parameter setting for the METABRIC dataset, we should maximize the homophily levels of our graph by trying different values for $k$ (KNN) and $r$ Radius. 

Since it can be noticed that for bigger homophily levels the accuracies also increase, we can conclude that the proposed models do separate the two classes of nodes for graph structures in which nodes with same class are favor an edge between them. 

Another conclusion which can be drawn from the diagrams is that, the models do work. For example for homophily levels of 74\%:
\begin{itemize}
    \item CNC-VGAE will produce representations that will return 90\% accuracy with a Naive Bayes classifier. 
    \item CNC-DGI will produce representations that attain 90\% accuracy with a Naive Bayes classifier. 
    \item 2G-DGI will produce representations that attain an accuracy of 90\%  with a Naive Bayes classifier
    \item Hetero-DGI will produce representations that attain an accuracy of 80\%  with a Naive Bayes classifier
    
\end{itemize}

Another aspect that can be observed is that for small homophily, less layers of GCN give better result. This can happen because as the number of GCN layers increases, nodes will get information from further neighbours, which might be of different label, so not representative for the for the original node's label.
\section{Evaluation on METABRIC}

This section, will introduce two evaluation procedures of the novel models on the METABRIC dataset. Since, the previous section has proved how increasing homophily levels also increase accuracies of the proposed models in the first evaluation experiment, we will carry experiments

\subsection{Graph hyper-parameter selection}
Since for pairs of modalities, and label class the construction of the graph is different the hyper-parameter search has been done on all pairs of modalities and all labels. What we have noticed is that the behavior of the results was constant through the modality change, but very different trough the label change class. Next, the results will be posted for all models on interaction of Clin+mRNA for all classes of labels. 

The tests we will carry vary the value of $k \in \{2, 4, 16, 64\}$ and $r \in \{0.005, 0.05, 0.5, 1, 5\}$. For all models the latent-lower space representation has 64 dimension. The dense layers have 128 dimensions. Each convolution has an PReLU (Parametric- ReLU) activation function. 
For each model in part the individual decisions we have took are:
\begin{itemize}
    \item For \textbf{CNC-VGAE} in special the reparametrisation function will be MMD, even thought the synthetic dataset the results where questionable. 
    \item For \textbf{2G-DGI} the special integration layer will be a simple average of the two lower representations, because on the synthetic dataset 
    \item For \textbf{Hetero-DGI} the special integration layer will be the dense layer
\end{itemize}

All the other tables can be find in the Appendix section.

\begin{figure}[!h]
    \small
    \centering
    \includegraphics[scale=0.38]{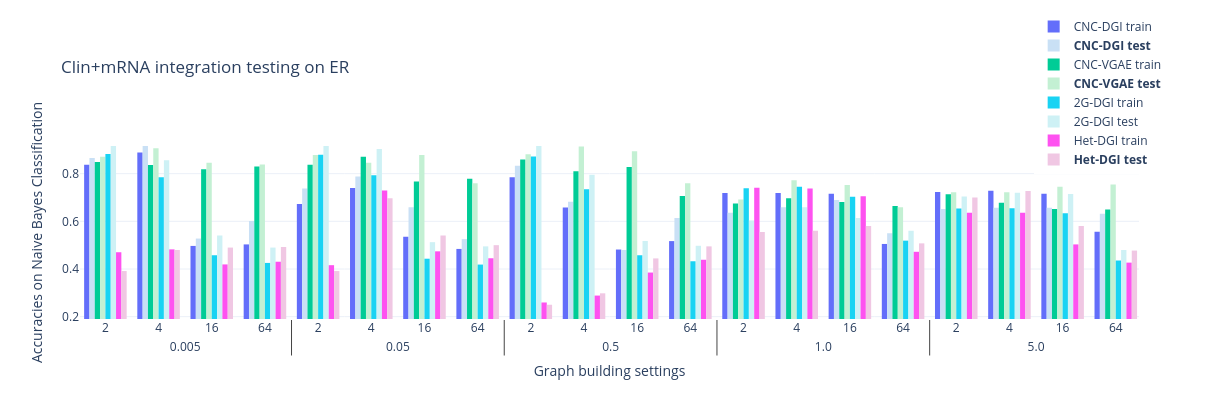}
    \caption{Accuracies of lower-latent representation obtained from CNC-VGAE}
    \label{fig:Clin+mRNA - ER}
\end{figure}
\begin{figure}[!h]
    \small
    \centering
    \includegraphics[scale=0.38]{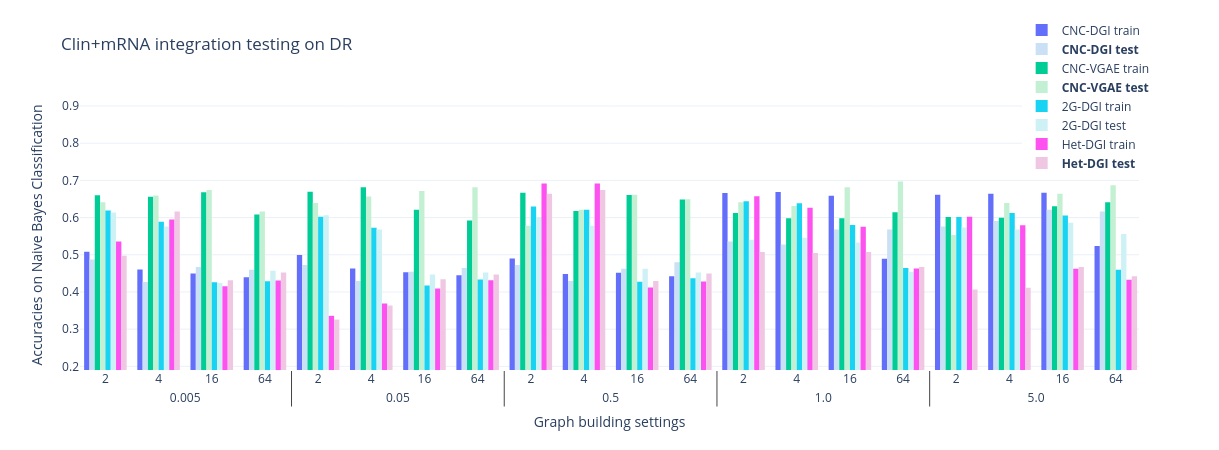}
    \caption{Accuracies of lower-latent representation obtained from CNC-VGAE}
    \label{fig:Clin+mRNA - DR}
\end{figure}
\begin{figure}[!h]
    \small
    \centering
    \includegraphics[scale=0.38]{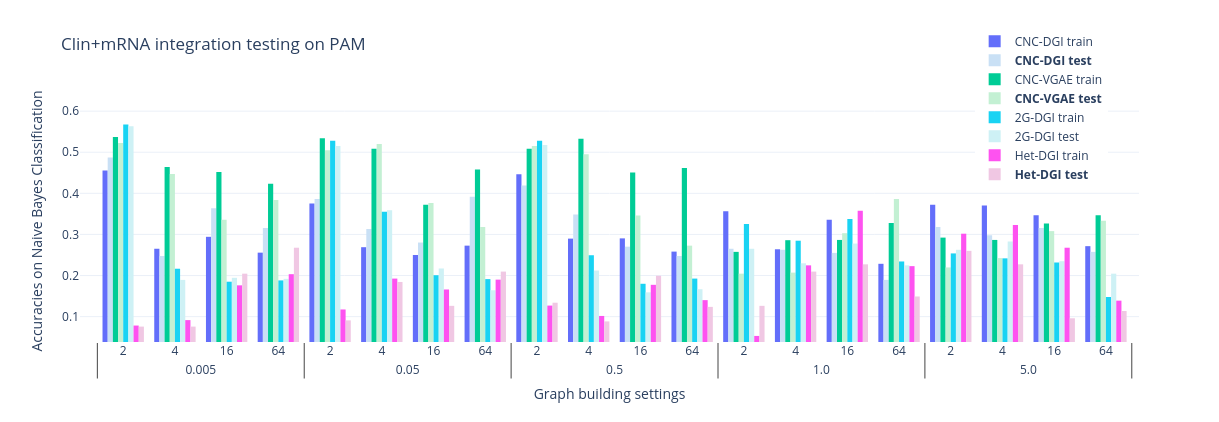}
    \caption{Accuracies of lower-latent representation obtained from CNC-VGAE}
    \label{fig:Clin+mRNA - PAM}
\end{figure}

\begin{figure}[!h]
    \small
    \centering
    \includegraphics[scale=0.38]{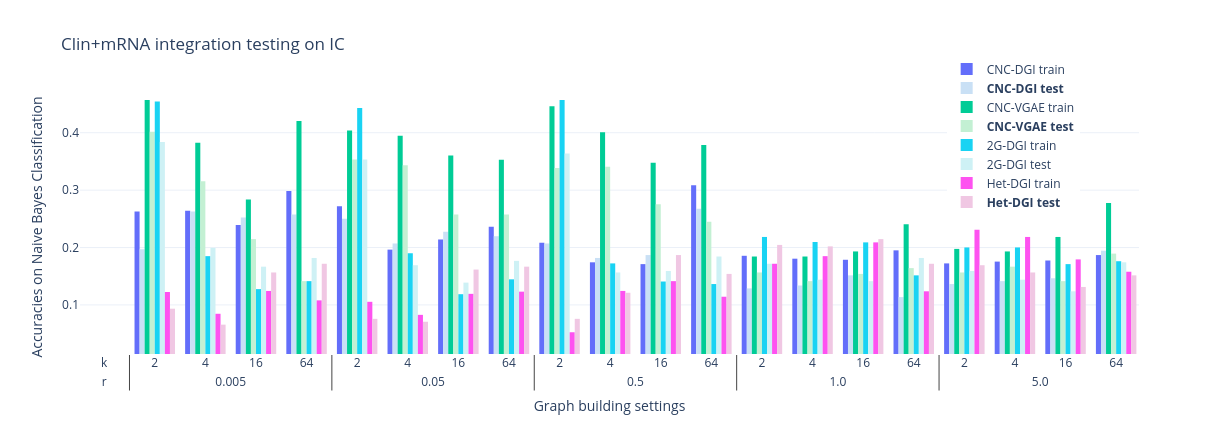}
    \caption{Accuracies of lower-latent representation obtained from CNC-VGAE}
    \label{fig:Clin+mRNA - IC}
\end{figure}
\clearpage

\subsubsection{Best Model Assessment}

\begin{table}[!h]
\centering
\resizebox{\textwidth}{!}{%
\begin{tabular}{cccccccccccccc}
\multicolumn{2}{c}{\multirow{2}{*}{}} & \multicolumn{3}{c}{CNC-DGI} & \multicolumn{3}{c}{CNC-VGAE} & \multicolumn{3}{c}{2G-DGI} & \multicolumn{3}{c}{Hetero-DGI} \\
\multicolumn{2}{c}{} & Clin+CNA & Clin+mRNA & CNA+mRNA & Clin+CNA & Clin+mRNA & CNA+mRNA & Clin+CNA & Clin+mRNA & CNA+mRNA & Clin+CNA & Clin+mRNA & CNA+mRNA \\
\multirow{3}{*}{ER} & NB & 0.712 & \textbf{0.939} & 0.712 & \textbf{0.835} & 0.914 & 0.835 & 0.694 & 0.927 & \textbf{0.924} & 0.758 & 0.727 & 0.755 \\
 & SVM & \textbf{0.793} & 0.919 & 0.881 & 0.841 & 0.914 & 0.851 & 0.773 & \textbf{0.929} & \textbf{0.939} & 0.763 & 0.770 & 0.768 \\
 & RF & 0.841 & \textbf{0.934} & 0.891 & 0.833 & 0.909 & 0.833 & 0.806 & 0.924 & \textbf{0.937} & 0.795 & 0.823 & 0.806 \\
\multirow{3}{*}{DR} & NB & 0.636 & 0.621 & 0.676 & 0.68 & 0.696 & 0.694 & 0.689 & 0.614 & 0.679 & 0.692 & 0.674 & 0.694 \\
 & SVM & 0.696 & 0.696 & 0.696 & 0.69 & 0.696 & 0.696 & 0.697 & 0.697 & \textbf{0.697} & 0.697 & 0.697 & \textbf{0.699} \\
 & RF & 0.703 & 0.699 & 0.694 & \textbf{0.71} & 0.704 & 0.699 & 0.697 & 0.705 & \textbf{0.717} & 0.677 & 0.674 & 0.672 \\
\multirow{3}{*}{PAM} & NB & 0.312 & 0.487 & 0.381 & 0.398 & 0.398 & 0.449 & \textbf{0.553} & \textbf{0.563} & 0.298 & 0.412 & 0.268 & 0.194 \\
 & SVM & 0.441 & 0.58 & 0.578 & 0.454 & 0.454 & 0.502 & \textbf{0.621} & 0.614 & 0.477 & 0.449 & 0.465 & 0.457 \\
 & RF & 0.497 & 0.58 & 0.563 & 0.457 & 0.457 & 0.515 & \textbf{0.644} & \textbf{0.652} & \textbf{0.576} & 0.422 & 0.518 & 0.452 \\
\multirow{3}{*}{IC} & NB & 0.391 & 0.267 & 0.31 & 0.497 & 0.401 & \textbf{0.520} & 0.447 & 0.384 & \textbf{0.538} & 0.260 & 0.215 & 0.215 \\
 & SVM & 0.458 & 0.387 & 0.454 & 0.482 & 0.414 & 0.532 & 0.467 & 0.482 & \textbf{0.593} & 0.391 & 0.278 & 0.338 \\
 & RF & 0.5 & 0.447 & 0.515 & 0.474 & 0.383 & \textbf{0.527} & 0.465 & 0.500 & \textbf{0.621} & 0.422 & 0.407 & 0.381
\end{tabular}%
}
\caption{Best-in-class results on representations obtained with the models trained on various settings on classification task obtained with Naive Bayes Classifier, Support Vector Machine and Random Forest}
\label{tab:best}
\end{table}
From Table \ref{tab:best} we can clearly understand that 2G-DGI obtains best-in-class results, compared to the other models. A nice surprise is that on Clin+CNA on PAM class it actually \textbf{beats the state of the art results, with $10\%$}.

Even though, the architectures of the 2G-DGI and Hetero-DGI where similar in some sense, there is a clear difference between the best results obtained by both of them. This will be investigated in future work.

General unsatisfying results on IC label class, can be motivated by the low homophily levels of the produced graph on this label class (between 17\%-19\%).
\subsubsection{Conclusions}

From the above Figures (\ref{fig:Clin+mRNA - ER}, \ref{fig:Clin+mRNA - DR}, \ref{fig:Clin+mRNA - PAM}, \ref{fig:Clin+mRNA - IC}) we can learn the followings, individual comparisons per model: 
\begin{itemize}
    \item \textbf{CNC-VGAE} must give the best results out of all models in this testing settings, reason for which we will test the values it will return for the KL reparametrisation loss. For low values of r and k it returns best accuracies. For all classes of labels there seem to be a jump in average when transitioning from $r=0.5$ to $r=1$. Generally the difference between the testing accuracy and the training accuracy are small, sometimes testing accuracies are higher than the training ones. 
    \item \textbf{Hetero-DGI} gives generally worse results than all the other models, this might be because the \textit{special integration layer} is a dense layer, and not an average one. Also one can notice generally pick for high values of r, rather than changes in k, in fact it seems to decrease as k grows. 
    \item \textbf{CNC-DGI} gives good results for the ER label, which has better homophily levels, where we can notice the average decreases as k increases. For DR label class, generally it gives good results when both $r$ and $k$ increase in value
    \item \textbf{2G-DGI} returned competitive results with CNC-VGAE, which was a nice surprise. On ER the highest test accuracy is 87\%, on DR is 69\%, on PAM 57\% (best out of all of them). 
\end{itemize}

Specific on the label classes, we can learn the following:
\begin{itemize}
    \item On \textbf{ER}, the models will return generally averages above 70\%
    \item On \textbf{DR} for big values of both $r$ and $k$ the results will generally be above 60\%
    \item On \textbf{PAM} most models return small accuracies (bellow 40\%) with the exception on CNC-VGAE and 2G-DGI which will get to accuracies of 55\% for values of $r$ smaller than $0.5$
    \item On \textbf{IC} most models will return small accuracies bellow 20\%, but for small graphs 2G-DGI and CNC-VGAE can get to accuracies of 40\%.
\end{itemize}

\begin{figure}[h]
    \small
    \centering
    \includegraphics[scale=0.35]{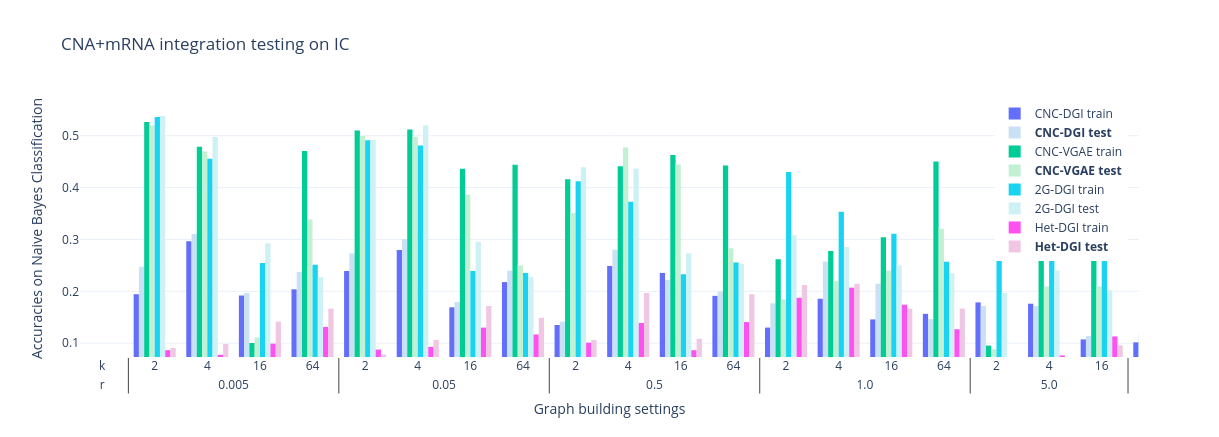}
\caption{CNA+mRNA integration testing on IC}
\label{fig:cna+mrna ic}
\end{figure}

Generally, from the above conclusions we can learn, that there exist some correlation between the homophily levels described in Tables \ref{tab:clinK},\ref{tab:rnaK},\ref{tab:cnaK}, \ref{tab:clinR}, \ref{tab:rnaR}, \ref{tab:cnaR} the number of edges in the graph, and the accuracies obtained. For \textbf{Clin}, homophily levels where around \textbf{16\%}, so this is a reason why IC on mRNA+Clin returns such small results. This exact fact can be proved by looking at Figure \ref{fig:cna+mrna ic}, which returns a best result of 53\% accuracy on IC for CNA+mRNA integration. Intuitively, it's almost like our learning process is downgraded by the high level of intra-class edges.

\chapter{Conclusion}
\section{Summary}
This project presents the reader with a deep dive into a novel unsupervised learning pipeline leveraging Graph Neural Networks on a cancer classification task. We commenced by discussing and recreating the state-of-the-art models in ``Variational Autoencoders for Cancer Data Integration: Design Principles and Computational Practice" \cite{SimidjievskiEtAl2019}, namely \textbf{CNC-VAE} and \textbf{H-VAE}. Our implementation of these architectures trained on the METABRIC dataset obtained results in line with the paper, and it provides a benchmark for the novel graph models proposed in our work.
\begin{itemize}
    \item The integration of Clin+mRNA on the IC label class with CNC-VAE rendered $79\%$ accuracy
    \item The integration of CNA+mRNA and Clin+mRNA on the PAM label class resulted in $68\%$, and $73\%$ respective accuracies
    \item On all integration types of the ER label class, accuracies were above $85\%$
\end{itemize}

The following topic focused on graph construction algorithms on data sets which do not store relations between the data points representing our patients. These approaches include KNN, and generating links based on the Euclidean distance between nodes. We defined metrics quantifying the characteristics and overall quality of such graph data sets, such as homophily, and analysed the resulted graphs using these measurements. Generally, lower homophily levels resulted in very low accuracies in the lower-latent representation evaluation phase, while high levels of homophily achieved up to 99.8\% accuracy. We can infer that the proposed models are sensitive to the graph structure of the input data.

During the design phase of the integrative models, we considered many factors such as the shape of the \textit{special integration layer} being parametric or non-parametric, number of layers in the autoencoders as well as the number of neurons in each layer and many others. Hyper-parameter fine-tuning has been performed on each model for all pairs of modalities and for each class of labels, and the evaluation process has been in line with the that used in state-of-the-art works in order to ensure consistency.

To prove the functionality of the novel models, we introduced a synthetic dataset for which the results observed using generated lower-dimensional embeddings on classification tasks with Naive Bayes vary between 51\% and 98\% accuracy. Specifically, on each homophily class:
\begin{itemize}
\item For homophily level of 51\%, 2G-DGI returned an accuracy of 82\%, and CNC-DGI returned 80\%.
\item For homophily level of 61\%, 2G-DGI returned an accuracy of 84\%, and Hetero-DGI returned 79\% accuracy.
\item For higher homophily levels, we notice best-in-class results that are above 90\%
\end{itemize}

Finally, as for the graph models applied on the METABRIC dataset, results vary much depending on the integrated modalities and on the label class, from 17\% to 92\%, in direct correlation with the homophily values of each label class. From Table \ref{tab:best} we can clearly understand that 2G-DGI obtains best-in-class results, compared to the other models. And we can also notice that on Clin+CNA integration on PAM with class it actually \textbf{beats the state of the art results, with $10\%$}. 

\section{Further work}

During the development of the experiments in this projects and up to the report writing phase, we identified several opportunities to advance this line of research that could be tackled in the future. First, investing in the search of the most optimal hyper-parameters for the graph construction algorithms and Graph Neural Networks proposed in this paper can help improve the current results. Second, analysing and extending the number of integrated modalities with, for example, visual data, has the potential of discovering deeper insights into cancer sub-type cluster and cancer classification. Finally, we propose a novel model adapted from the Hierarchical Variational Autoencoder that introduces the use of Graph Convolutional Layers after the initial enconding phase. We conclude by presenting a mathematical problem regarding graph data that could be solved with probability theory and combinatorics.

\subsubsection{Hyper-parameter settings}
Given the multitude of architectural decisions required by the model training trials in this project, we intend to carry out more tests and search for the most optimal hyper-parameter settings that will further advance our architectures, such as the \textit{parameterised special integrative layer}, the depth of the  Encoder (i.e. number of GCN layers used), the size of the dense and latent space layers and so on.

\subsubsection{Multi-modal expansion}
The modalities integrated in this paper represent either continuous or categorical data - we intend to extend the integrative capabilities of the network to image data, and thus use more than two modalities. For the visual data, CNN layers can be added prior to the integration phase of the models to extract higher-level features from the input images.

\subsubsection{H-VGAE}
To advance the research avenue tackled in this project, we propose another model adapted from the Hierarchical Variational Autoencoder \cite{SimidjievskiEtAl2019}, which will be named Hierarchical Graph Variational Autoencoder (H-VGAE). The first processing units in this model are comprised of a series of autoencoders, that will generate a lower-latent representations independently for each input modality. A graph construction algorithm will be applied further to build relationships across the resulted embeddings, which will be fed among the lower-dimensional representations to two Graph Convolutional Layers: one aiming to find the mean and one to find the variance of the encoding distribution. Finally, the decoding phase consists of an inner product operation on the final representation, which will be compared to the originally built graph in the loss function.

\begin{figure}[h]
    \small
    \centering
    \includegraphics[scale=0.40]{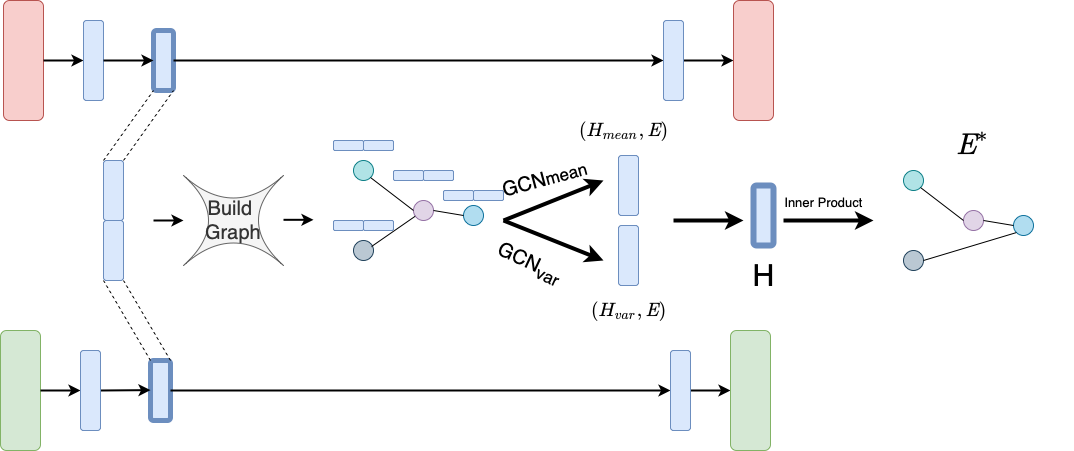}
    \caption{\textbf{H-VGAE} proposed architecture for integration}
    \label{fig:vae}
\end{figure}

The reasons for which this model has the potential to render competitive results:
\begin{itemize}
    \item The lower-dimensional embeddings generated by the first layer of autoencoders (one for each input modality) will lie on a continuous multi-Gaussian space. Hence, the radius algorithm for generating edges has a higher probability of returning dense graphs with better homophily levels than by using the method on its own.
    \item The closes model to this new architecture, CNC-VGAE, obtained among the best results across all tested models.
\end{itemize}
\subsubsection{A Math Problem}
\begin{figure}[h]
    \small
    \centering
    \includegraphics[scale=0.30]{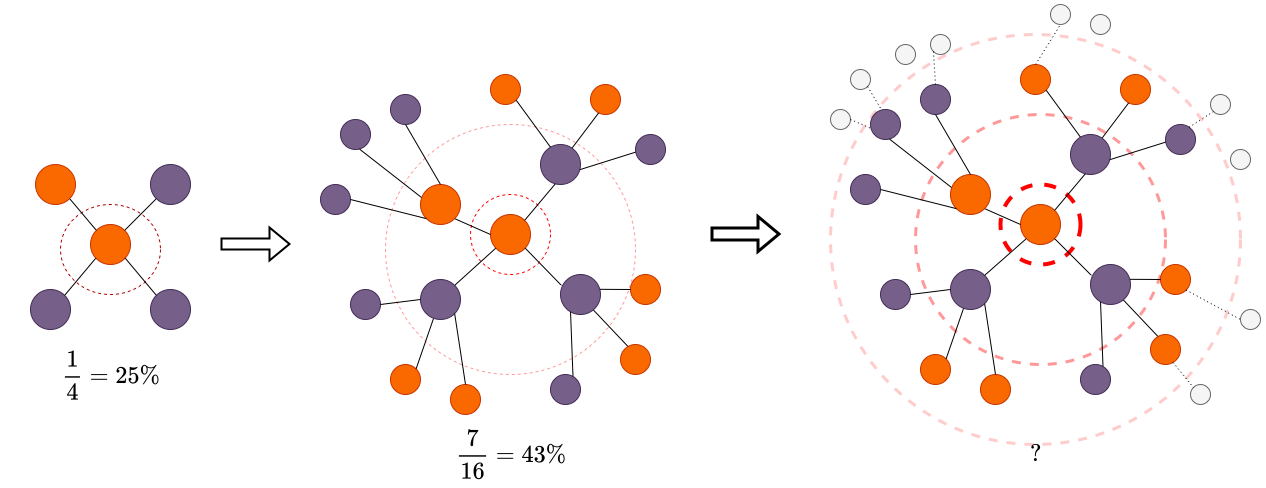}
    \caption{Growing neighbourhoods, and growing number of like neighbours}
    \label{fig:maths_problem}
\end{figure}
Take a graph with $25\%$ homophily level. Let O (orange) and P (purple) be two labels that the nodes can take, and let's pick a node of label O. By taking the nodes immediate neighbourhood we expect that 1 out of 4 neighbours to be of label O, conversely for each P (purple) node we expect 3 orange neighbours and 1 purple neighbour. By taking a bigger neighbour, that includes immediate neighbours and their neighbours, we expect that 4 out of 10 to be of label O. This can be clearly understood from Figure \ref{fig:maths_problem}. Our open ended question is if we continue increasing the neighborhoods can we reach a maximum for same label neighbours, will the number converge? Does this happen for classes that have more than two labels? What about different homophily levels? Can we generalize a formula?

This question can be relevant in this context because the number of growing nested neighbourhoods, can be the number of convolution layers that we apply, to a dataset with a certain homophily level. Attempting to answer this question might raise ideas on how learning on graphs with small homophily levels should be attempted.

\bibliography{mwe}       
\bibliographystyle{alpha}       
\chapter{Appendix}
\section{Supplementary results}
\begin{figure}[h]
    \small
    \centering
    \includegraphics[scale=0.35]{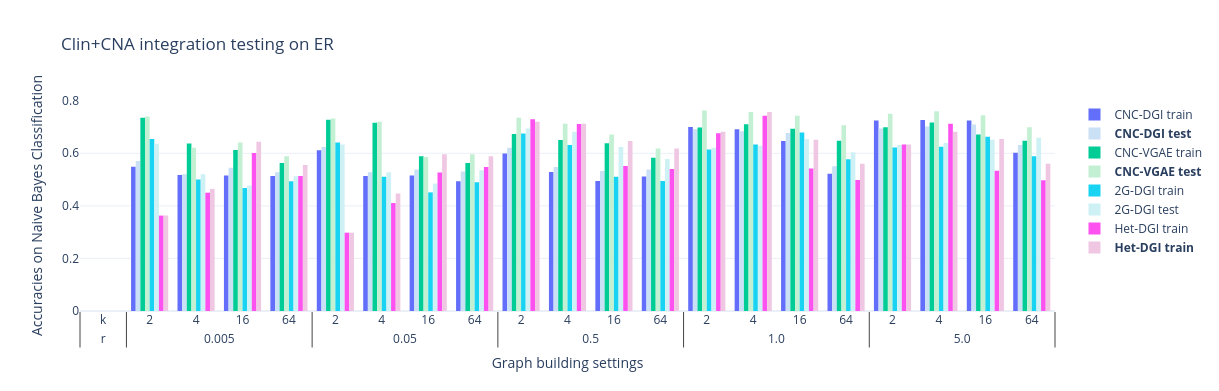}
\caption{Clin+CNA integration testing on ER}
\end{figure}
\begin{figure}[h]
    \small
    \centering
    \includegraphics[scale=0.35]{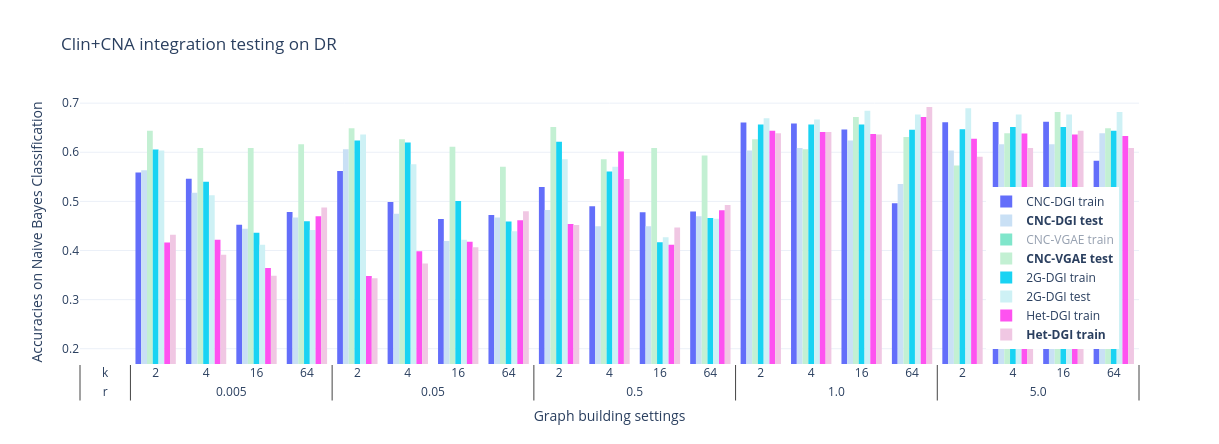}
\caption{Clin+CNA integration testing on DR}
\end{figure}
\begin{figure}[h]
    \small
    \centering
    \includegraphics[scale=0.35]{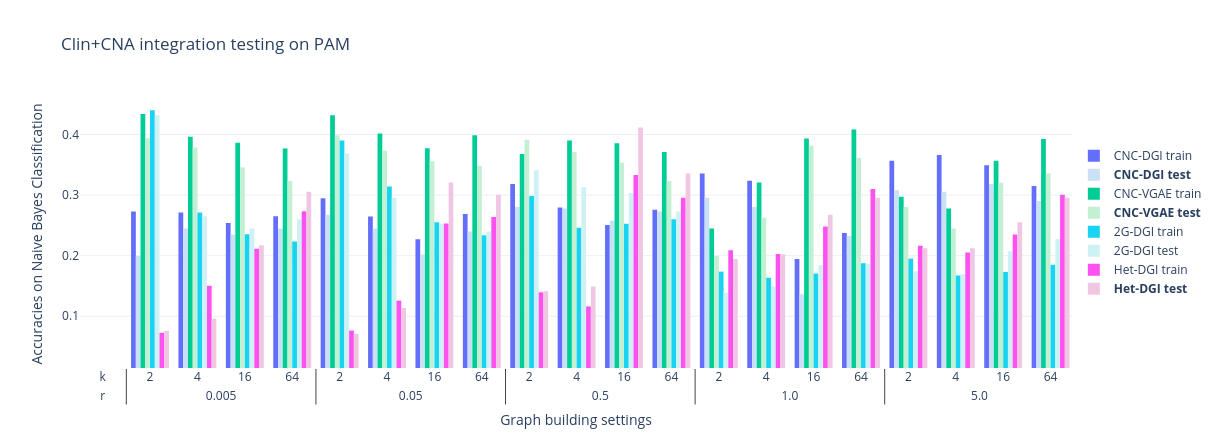}
\caption{Clin+CNA integration testing on PAM}
\end{figure}
\begin{figure}[h]
    \small
    \centering
    \includegraphics[scale=0.35]{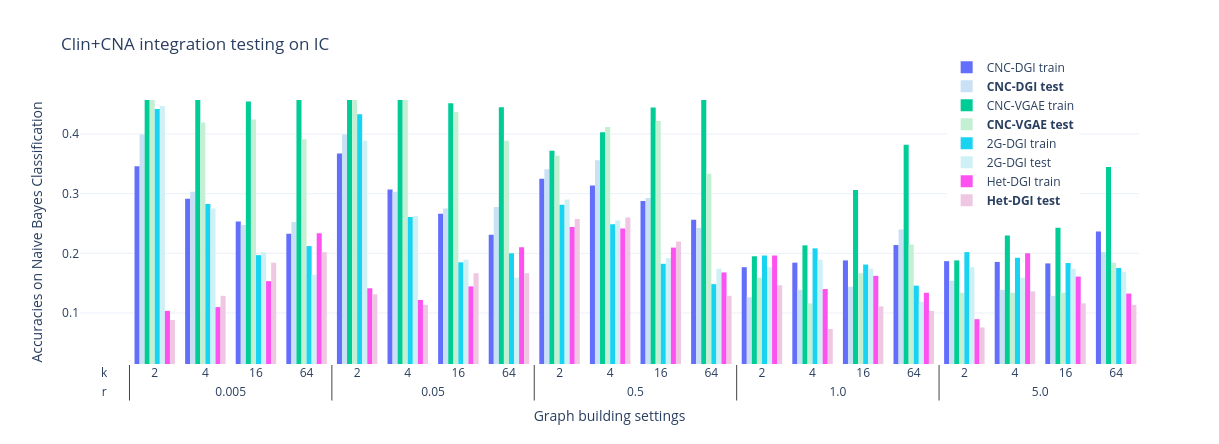}
\caption{Clin+CNA integration testing on IC}
\end{figure}
\begin{figure}[h]
    \small
    \centering
    \includegraphics[scale=0.35]{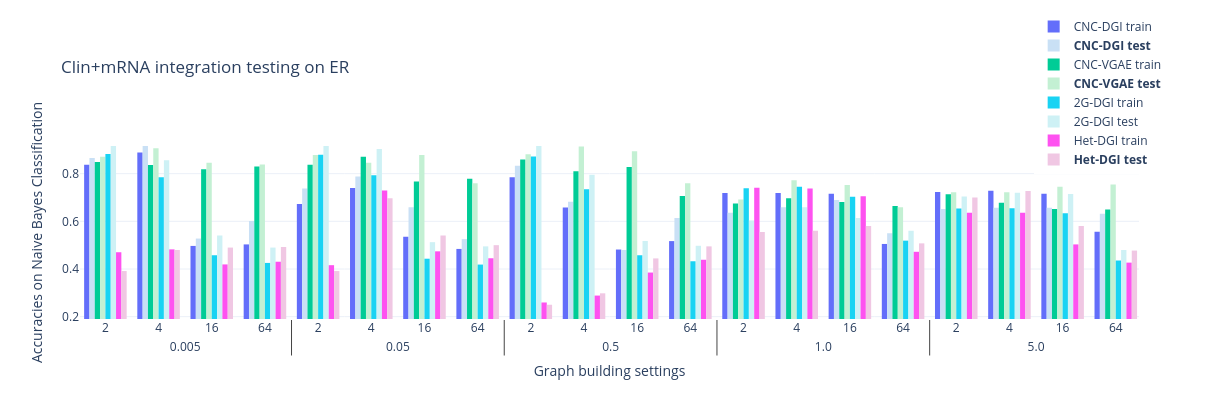}
\caption{Clin+mRNA integration testing on ER}
\end{figure}
\begin{figure}[h]
    \small
    \centering
    \includegraphics[scale=0.35]{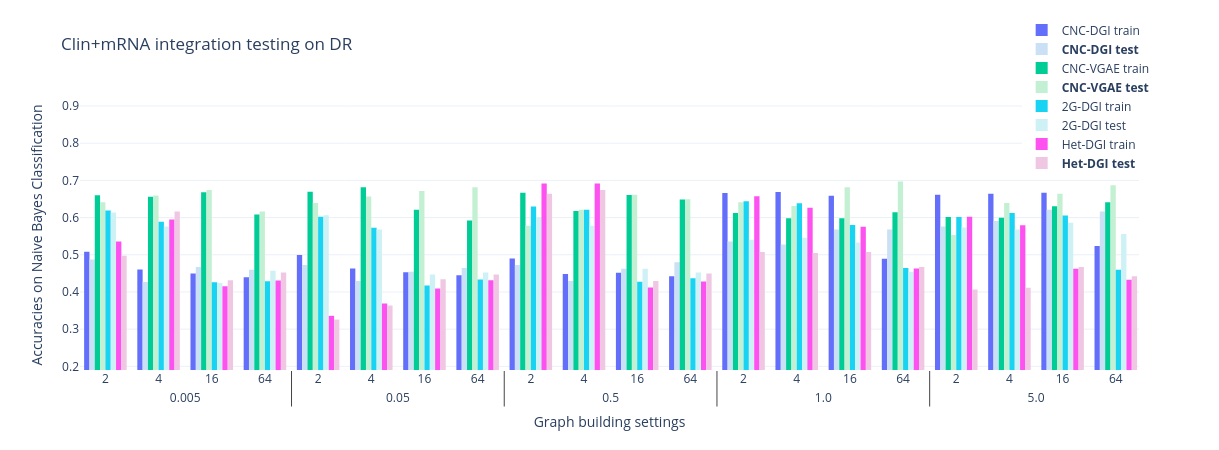}
\caption{Clin+mRNA integration testing on DR}
\end{figure}
\begin{figure}[h]
    \small
    \centering
    \includegraphics[scale=0.35]{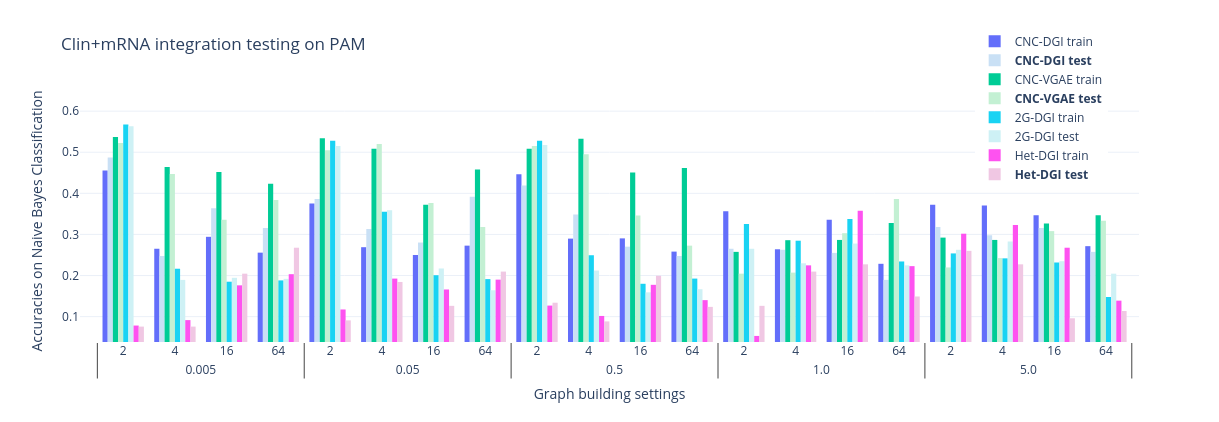}
\caption{Clin+mRNA integration testing on PAM}
\end{figure}
\begin{figure}[h]
    \small
    \centering
    \includegraphics[scale=0.35]{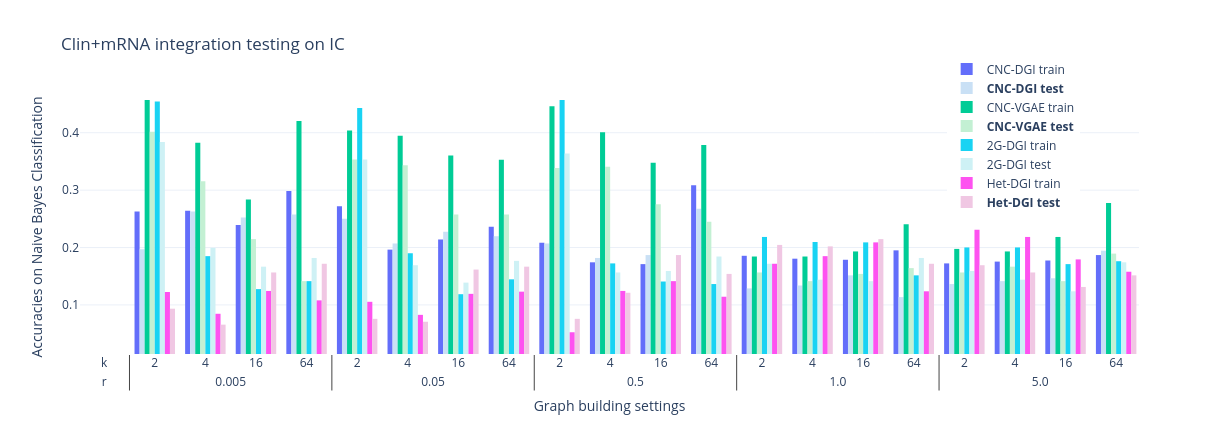}
\caption{Clin+mRNA integration testing on IC}
\end{figure}
\begin{figure}[h]
    \small
    \centering
    \includegraphics[scale=0.35]{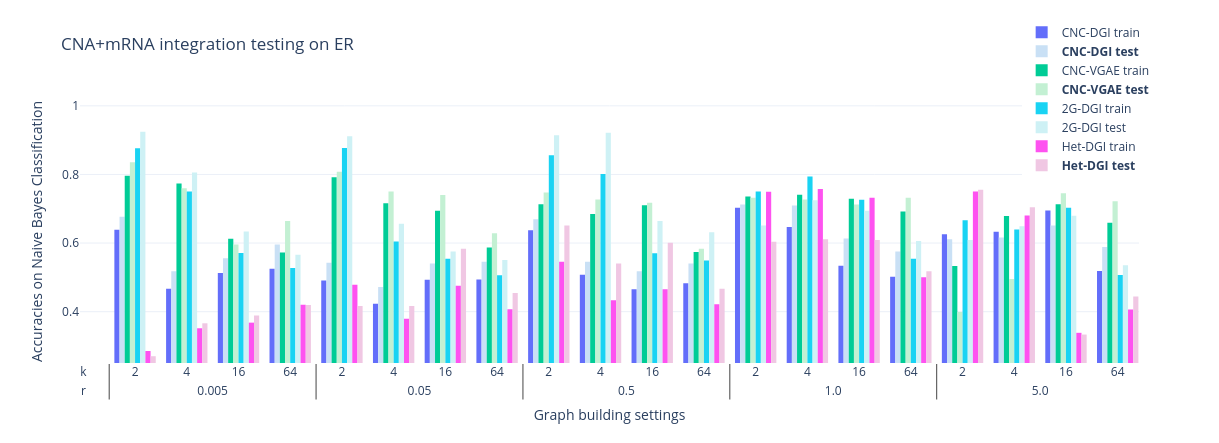}
\caption{CNA+mRNA integration testing on ER}
\end{figure}
\begin{figure}[h]
    \small
    \centering
    \includegraphics[scale=0.35]{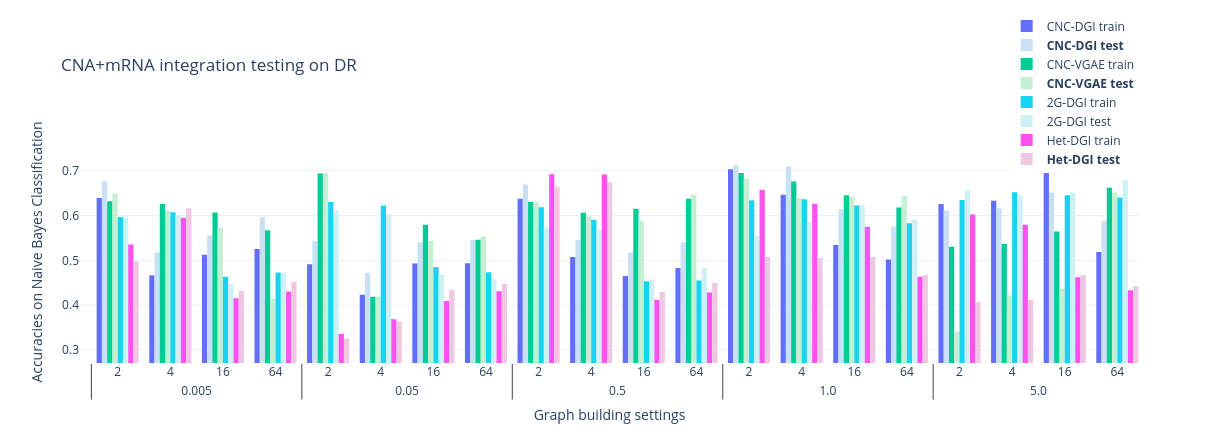}
\caption{CNA+mRNA integration testing on DR}
\end{figure}
\begin{figure}[h]
    \small
    \centering
    \includegraphics[scale=0.35]{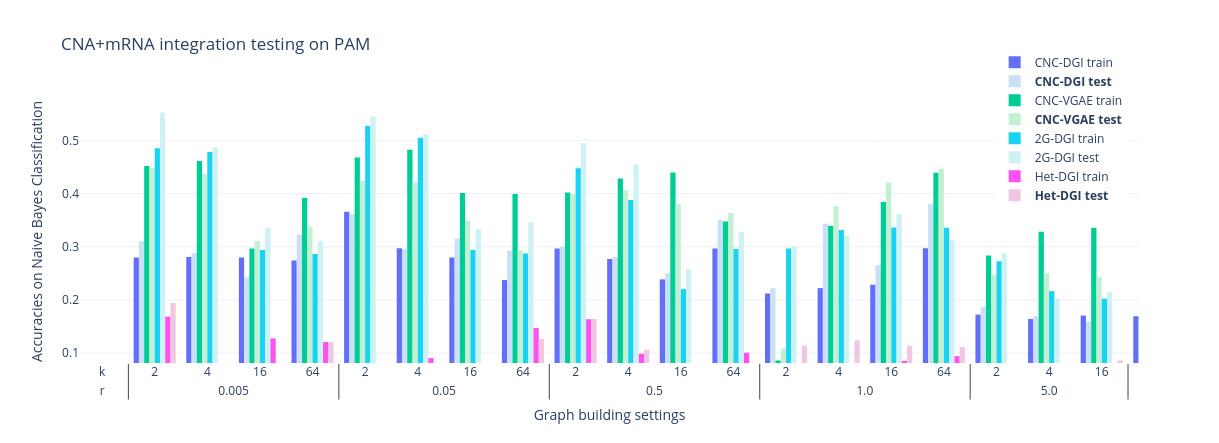}
\caption{CNA+mRNA integration testing on PAM}
\end{figure}
\end{document}